\pdfoutput=1

\documentclass[11pt]{article}

\usepackage{EMNLP2022}

\usepackage{times}
\usepackage{latexsym}

\usepackage[T1]{fontenc}

\usepackage[utf8]{inputenc}

\usepackage{microtype}

\usepackage{inconsolata}

\usepackage{hyperref}
\usepackage{xcolor}
\usepackage{tabularx}
\usepackage{graphicx}
\usepackage{booktabs}

\usepackage{amsmath}
\usepackage{multirow}
\usepackage{subcaption}

\usepackage{amssymb}
\usepackage{colortbl}
\usepackage{arydshln}
\usepackage{color}

\title{Late Prompt Tuning: A Late Prompt Could Be Better Than Many Prompts}

\author{
Xiangyang Liu\textsuperscript{\rm 1,2} \quad Tianxiang Sun\textsuperscript{\rm 1,2} \quad Xuanjing Huang\textsuperscript{\rm 1,2} \quad Xipeng Qiu\textsuperscript{\rm 1,2 \thanks{{} {} Corresponding author.}}
\\
\textsuperscript{\rm 1}School of Computer Science, Fudan University\\
\textsuperscript{\rm 2}Shanghai Key Laboratory of Intelligent Information Processing, Fudan University\\
\texttt{\{xiangyangliu20,txsun19,xjhuang,xpqiu\}@fudan.edu.cn}\\
}

\begin{document}
\maketitle
\begin{abstract}
Prompt tuning is a parameter-efficient tuning (PETuning) method for utilizing pre-trained models (PTMs) that simply prepends a soft prompt to the input and only optimizes the prompt to adapt PTMs to downstream tasks. Although it is parameter- and deployment-efficient, its performance still lags behind other state-of-the-art PETuning methods. Besides, the training cost of prompt tuning is not significantly reduced due to the back-propagation through the entire model.
Through empirical analyses, we shed some light on the lagging performance of prompt tuning and recognize a trade-off between the propagation distance from label signals to the inserted prompt and the influence of the prompt on model outputs. 
Further, we present \textbf{L}ate \textbf{P}rompt \textbf{T}uning (LPT) that inserts a late prompt into an intermediate layer of the PTM instead of the input layer or all layers.
The late prompt is obtained by a neural prompt generator conditioned on the hidden states before the prompt insertion layer and therefore is instance-dependent.
Through extensive experimental results across various tasks and PTMs, we show that LPT can achieve competitive performance to full model tuning and other PETuning methods under both full-data and few-shot scenarios while possessing faster training speed and lower memory cost.        

\end{abstract}

\section{Introduction}

Pre-trained models~\citep{devlin2019bert,radford2019gpt2,yang2019xlnet,raffel2020t5,lewis2020bart,liu2021elue,Qiu2020survey,survey2021lin} have pushed most NLP tasks to state-of-the-art. Model tuning (or fine-tuning) is a popular method for utilizing PTMs on downstream tasks that needs to tune all parameters of PTMs for every task. Despite the welcome outcome, it leads to prohibitive adaptation costs, especially for supersized PTMs~\citep{brown2020gpt3,ernie32021wang}. Parameter-efficient tuning (PETuning) is a new tuning paradigm that can adapt PTMs to downstream tasks by only tuning a very small number of internal or additional parameters. 

\begin{figure}[t!]
    \centering
    \includegraphics[width=\linewidth]{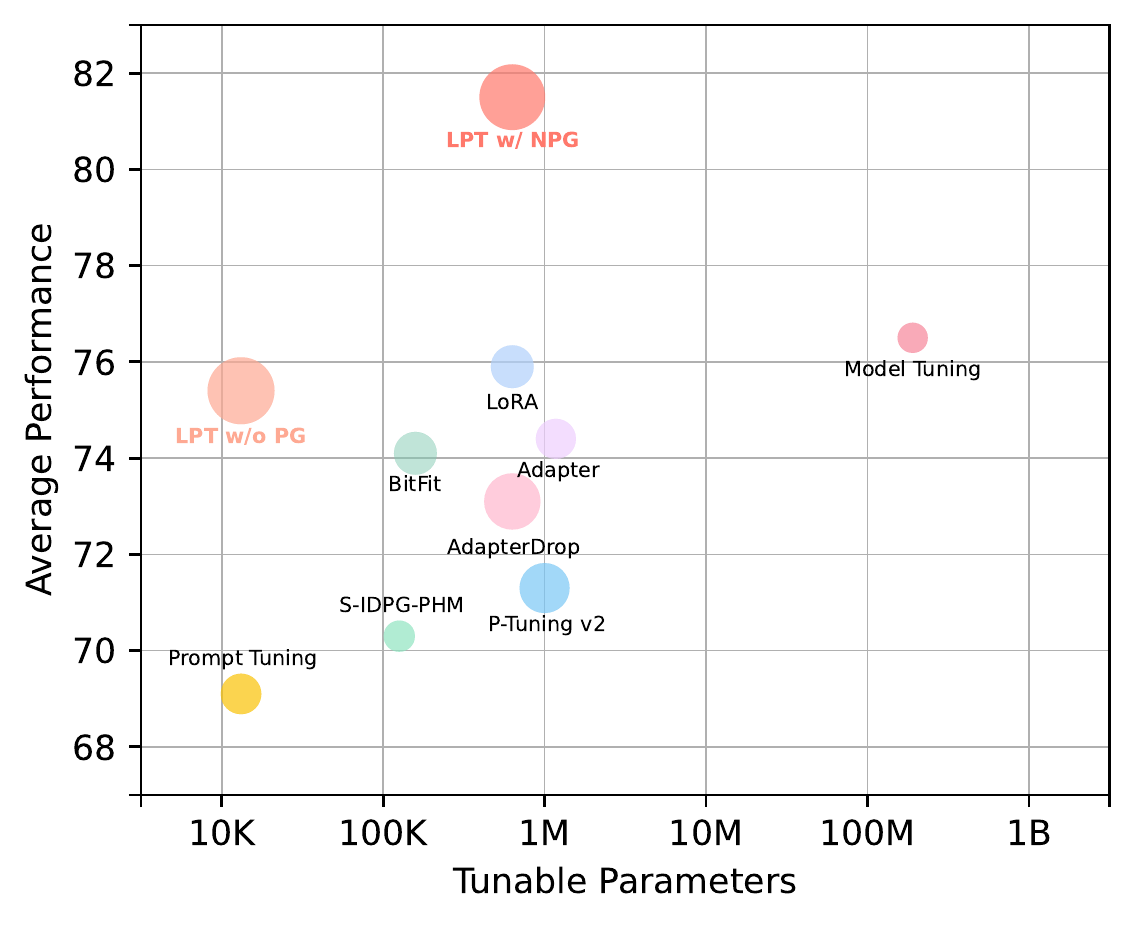}
    \caption{Overall comparison between LPT and baselines of only 100 training samples for each task. All methods are evaluated on 10 text classification tasks  using RoBERTa\textsubscript{LARGE}. The radius of every circle indicates training speed (tokens per millisecond). LPT w/ NPG and LPT w/o PG represent LPT with naive prompt generator and without prompt generator, respectively. The details can be found in Section~\ref{sec:lip}.}
    \label{fig:overall_comparison}
\end{figure}

Prompt tuning~\citep{lester2021pt} is a simple and popular PETuning method that prepends a sequence of soft prompt tokens to the input and only optimizes the prompt to adapt PTMs to downstream tasks. It has an absolute advantage in parameter efficiency and facilitates mixed-task inference, which makes the deployment of PTMs convenient. However, compared with other advanced PETuning methods, e.g., Adapter~\citep{Houlsby2019adapter,Mahabadi2021compacter}, LoRA~\citep{hu2022lora}, and BitFit~\citep{Zaken2022bitfit}, prompt tuning suffers from lower performance and convergence rate. Compared with full model tuning, although the number of trainable parameters in prompt tuning reduces by $\sim$17,000$\times$ (from 355M to 21K on RoBERTa\textsubscript{LARGE}), the training speed only increases by $\sim$1.5$\times$, and the memory cost only reduces by 29.8\%.\footnote{Refer to Section \ref{sec:efficiency} for details.} P-tuning v2~\citep{Liu2022ptuningv2} improves the performance of prompt tuning by inserting soft prompts into every hidden layer of PTMs, but it is difficult to optimize and needs more training steps to attain competitive performance. 

In this paper, we explore why prompt tuning performs poorly and find there is a trade-off between the propagation distance from label signals to the inserted prompt and the influence of the prompt on model outputs. The key to prompt tuning is to make the soft prompt carry task-related information through downstream training. The trained prompt can interact with text inputs during the model forward pass to obtain text representations with task-related information. Since the prompt is inserted into the input in prompt tuning, it has a strong ability to influence the outputs of PTM through sufficient interactions with text inputs. However, there is a long propagation path from label signals to the prompt. It leads us to ask the question: \textit{Does this long propagation path cause a lot of task-related information to be lost during propagation and thus affect performance?} To verify the impact of the propagation distance on performance, we conduct pilot experiments by shortening it in Section \ref{sec:late_prompt} and find that the performance first increases then decreases with the shortening of the length. This finding inspires us to present the \textbf{late prompt} (i.e., inserting the prompt into an intermediate hidden layer of PTM). The late prompt not only receives more task-related information at each update due to the shorter propagation path of task-related information but also maintains the adequate ability to influence the outputs of PTM. Despite the higher performance and faster convergence rate of late prompt than prompt tuning, the hidden states produced by PTM before the prompt insertion layer are underutilized. To further improve performance and take full advantage of these contextual hidden representations, we introduce a prompt generator to generate the soft prompt (termed as \textbf{instance-aware prompt}) for each instance using the corresponding hidden states.

Based on the late and instance-aware prompt, we present \textbf{L}ate \textbf{P}rompt \textbf{T}uning (LPT) to improve prompt tuning. Since the soft prompt is inserted into an intermediate layer of PTM, we have no need to compute gradients for model parameters below the prompt insertion layer, and therefore speed up the training process and reduce memory costs. Extensive experimental results show that LPT outperforms most prompt-based tuning methods and can be comparable with adapter-based tuning methods and even full model tuning. Especially in the few-shot scenario with only 100 training samples, LPT outperforms prompt tuning by \textbf{12.4 points} and model tuning by \textbf{5.0 points} in the average performance of ten text classification tasks. Besides, it is \textbf{2.0}$\times$ faster and reduces by \textbf{56.6\%} than model tuning in terms of training speed and memory cost on RoBERTa\textsubscript{LARGE}, respectively. Figure~\ref{fig:overall_comparison} shows an overall comparison between LPT and its counterparts. To sum up, the key contributions of this paper are:

\begin{itemize}
    \item We explore why prompt tuning performs poorly and find that it is due to the long propagation path from label signals to the input prompt and present a simple variant named late prompt tuning to address the issue.
    \item Combining the late and instance-aware prompts, we present LPT, which not only attains comparable performance with adapter-based tuning methods and even model tuning but also greatly reduces training costs.
    \item We verify the versatility of LPT in the full-data and few-shot scenarios across 10 text classification tasks and 3 PTMs. Code is publicly available at \url{https://github.com/xyltt/LPT}.
\end{itemize}

\section{Related Work}

\paragraph{Adapter-based tuning.} One research line of PETuning is adapter-based tuning~\citep{Ding2022deltatuning} that inserts some adapter modules between model layers and optimizes these adapters in downstream training for model adaptation. Adapter~\citep{Houlsby2019adapter} inserts adapter modules with bottleneck architecture between every consecutive Transformer~\citep{Vaswani2017transformer} sub-layers. AdapterDrop~\citep{Andreas2020adapterdrop} investigates the efficiency through removing adapters from lower layers. Compacter~\citep{Mahabadi2021compacter} uses low-rank optimization and parameterized hypercomplex multiplication~\citep{zhang2021phm} to compress adapters.
Adapter-based tuning methods have comparable results with model tuning when training data is sufficient but don't work well in the few-shot scenario~\citep{Wang2021list}.

\paragraph{Prompt-based tuning.} Another main research line of PETuning is prompt-based tuning that inserts some additional soft prompts into the hidden states instead of injecting new neural modules to PTMs. Prompt tuning~\citep{lester2021pt} and P-tuning~\citep{liu2021ptuning} insert a soft prompt to word embeddings only, and can achieve competitive results when applied to supersized PTMs. Prefix-tuning~\citep{Li2021prefixtuning} and P-tuning v2~\citep{Liu2022ptuningv2} insert prompts to every hidden layer of PTM. BBT~\citep{Sun2022bbt} optimizes the inserted prompt with derivative-free optimization. Some prompt-based tuning methods, like prompt tuning and BBT, formulate downstream tasks as pre-training tasks (e.g., masked language modeling task) to close the gap between pre-training and downstream training~\citep{sun2022paradigm}.
There are also some prompt-based methods with instance-aware prompt. IDPG~\citep{Wu2022idpg} uses the prompt generator with parameterized hypercomplex multiplication~\citep{zhang2021phm} to generate a soft prompt for every instance. Context-tuning~\citep{tang2022contexttuning} uses BERT model~\citep{devlin2019bert} as the prompt generator and focuses on NLG tasks. IPL~\citep{Jin2022ipl} first calculates relevance scores between prompt tokens and inputs, then uses the scores to re-weight the original prompt tokens. But it tunes all parameters of PTM. All the above methods with instance-aware prompt have the same weakness because they need to encode the inputs using an extra encoder, which slows down the training and increases inference latency.

There are also some other popular PETuning methods, such as BitFit~\citep{Zaken2022bitfit} which only tunes the bias terms, LoRA~\citep{hu2022lora} which optimizes low-rank decomposition matrices of the weights within self-attention layers.

\section{Problem Formulation}

Given a PTM $\mathcal{M}$, in the setting of model tuning, we first reformulate the inputs with single sentence as $\mathbf{E}$\texttt{([CLS] $\langle S_1 \rangle$ [SEP])} and the inputs with sentence pair as $\mathbf{E}$\texttt{([CLS]  $\langle S_1 \rangle$ [SEP] $\langle S_2 \rangle$ [SEP])}, where $\mathbf{E}$ is the embedding layer of $\mathcal{M}$. The final hidden state of \texttt{[CLS]} token will be used to predict label. In the setting of prompt tuning, we insert a randomly initialized soft prompt \textbf{p} into word embeddings, and also modify the original inputs using different manual templates with a \texttt{[MASK]} token for different tasks. For example, the inputs with single sentence from a sentiment analysis task will be transform into \texttt{concat(}\textbf{p}, $\mathbf{E}$\texttt{([CLS] $\langle S_1 \rangle$} It was \texttt{[MASK]. [SEP]))}. Then, we map the original labels $\mathcal{Y}$ to some words in the vocabulary $\mathcal{V}$ of $\mathcal{M}$, which formulates downstream tasks as a language modeling task to close the gap between pre-training and downstream training. The final hidden state of \texttt{[MASK]} token will be used to predict label. 

In the setting of our proposed method LPT, we use a prompt generator (\textbf{PG}) to generate an independent prompt \textbf{p} for every input. In addition, the layer that the prompt inserts into is an intermediate layer of PTM instead of word embeddings, and we refer to the layer as the prompt layer (\textbf{PL}).

\section{Why Prompt Tuning Performs Poorly?}
\label{sec:late_prompt}

The workflow of prompt tuning is to make the inserted soft prompt carry task-related information through downstream training. In the inference phase, this prompt can interact with test inputs during layer-upon-layer propagation so that the hidden representations of these inputs also contain task-related information. There are strong interactions between the prompt and text inputs because prompt tuning inserts prompt into word embeddings. However, there is a long propagation path from label signals to the prompt. Therefore, we speculate that the poor performance of prompt tuning is due to the long propagation path of task-related information, which causes a lot of task-related information to be lost during propagation in the frozen model and thus affect performance. To verify this conjecture, we conduct some pilot experiments on TREC~\citep{voorhees2000trec} and RTE~\citep{dagan2005rte} datasets using RoBERTa\textsubscript{LARGE}~\citep{liu2019roberta}.

\begin{figure*}[t!]
    \centering
    \begin{subfigure}{0.32\linewidth}
    \centering
    \includegraphics[width=\linewidth]{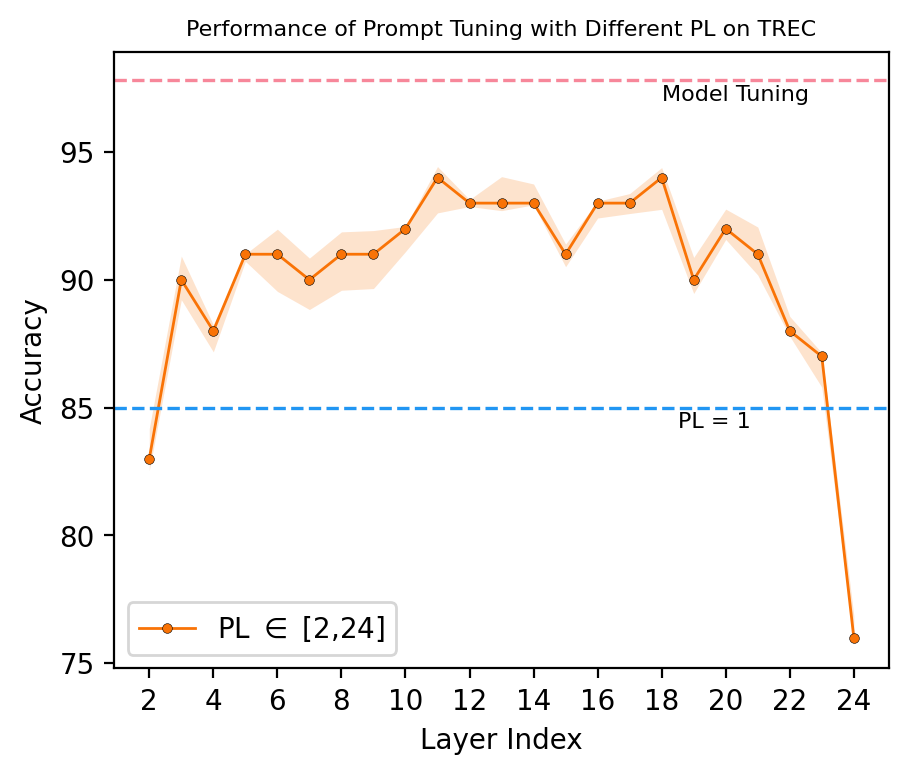}
    \end{subfigure}
    \begin{subfigure}{0.32\linewidth}
    \centering
    \includegraphics[width=\linewidth]{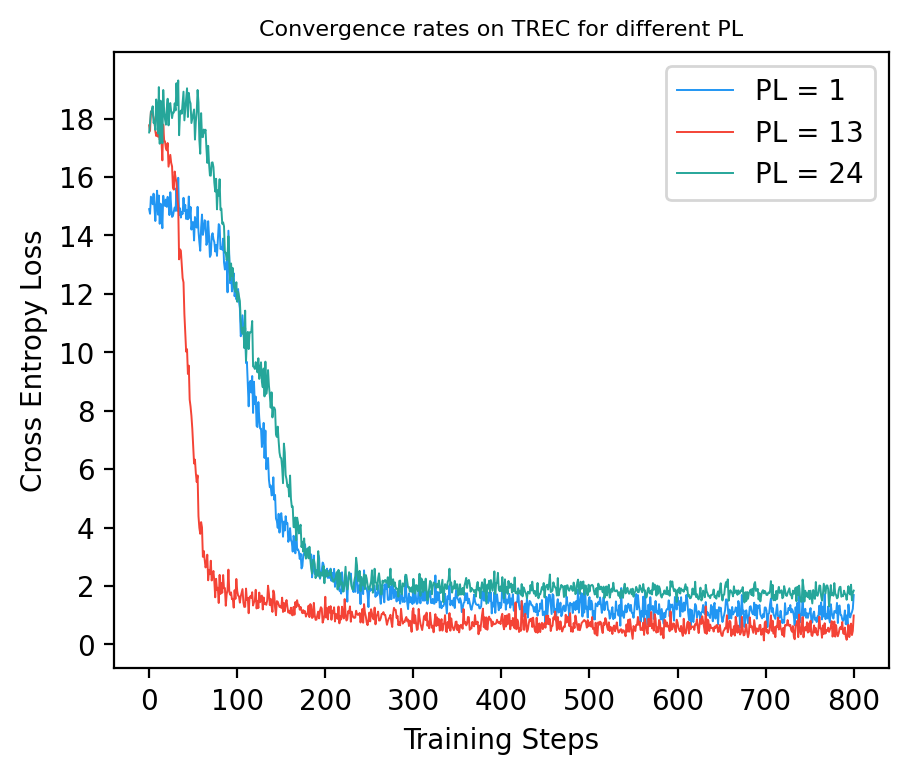}
    \end{subfigure}
    \begin{subfigure}{0.32\linewidth}
    \centering
    \includegraphics[width=\linewidth]{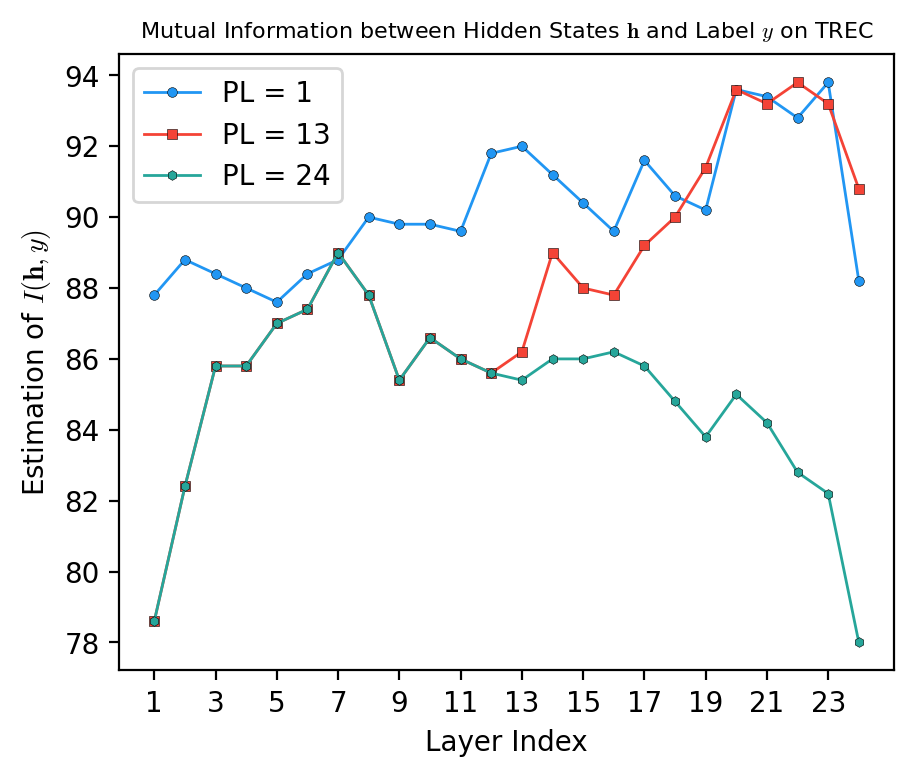}
    \end{subfigure}
    \\
    \begin{subfigure}{0.32\linewidth}
    \centering
    \includegraphics[width=\linewidth]{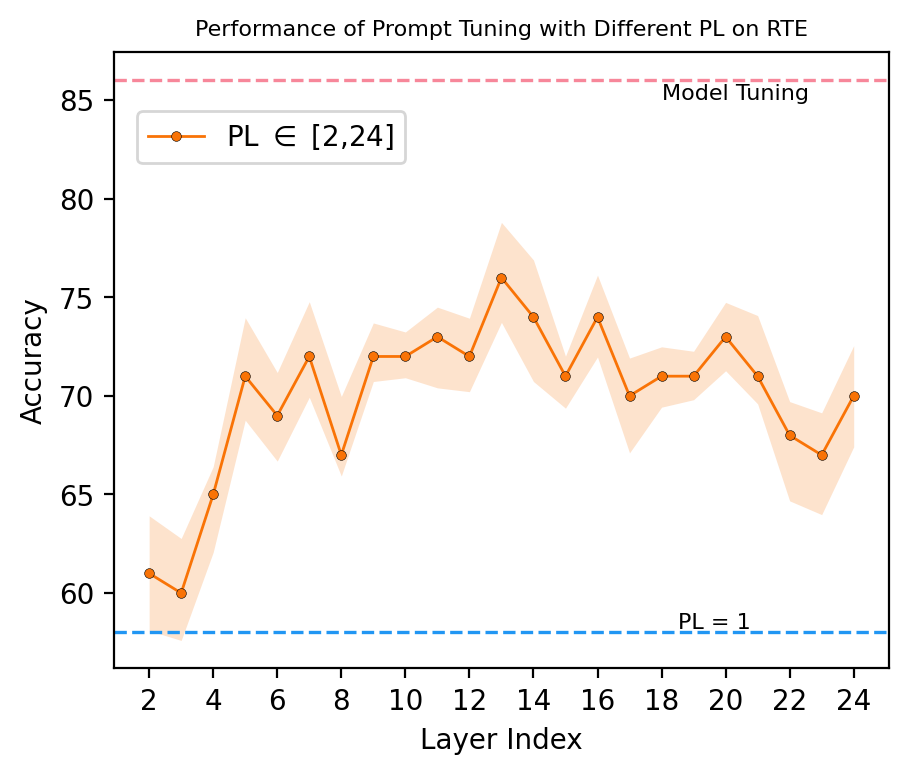}
    \end{subfigure}
    \begin{subfigure}{0.32\linewidth}
    \centering
    \includegraphics[width=\linewidth]{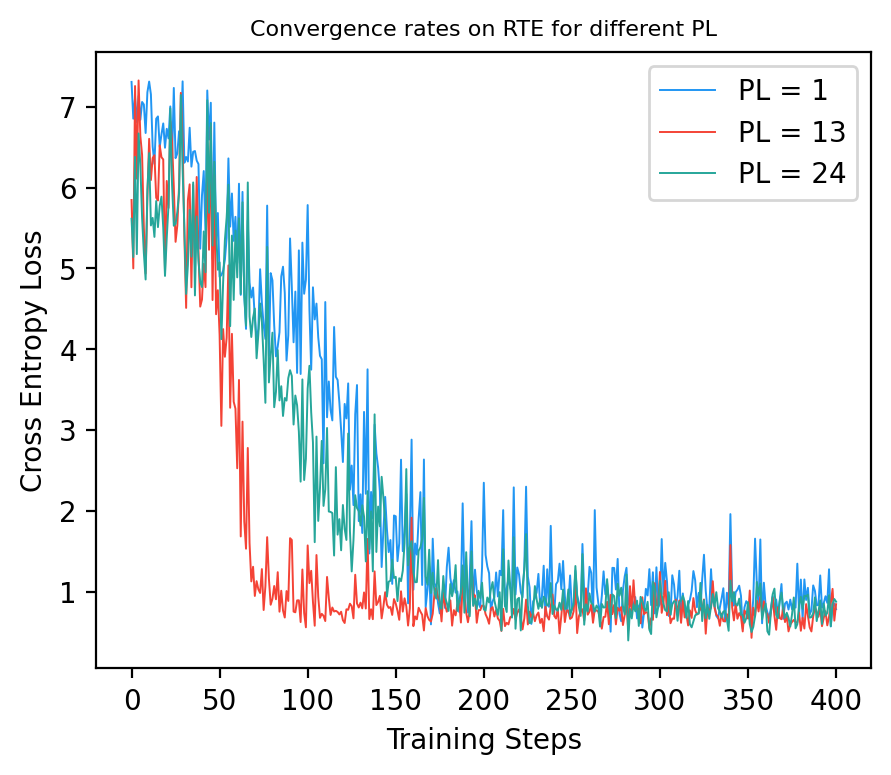}
    \end{subfigure}
    \begin{subfigure}{0.32\linewidth}
    \centering
    \includegraphics[width=\linewidth]{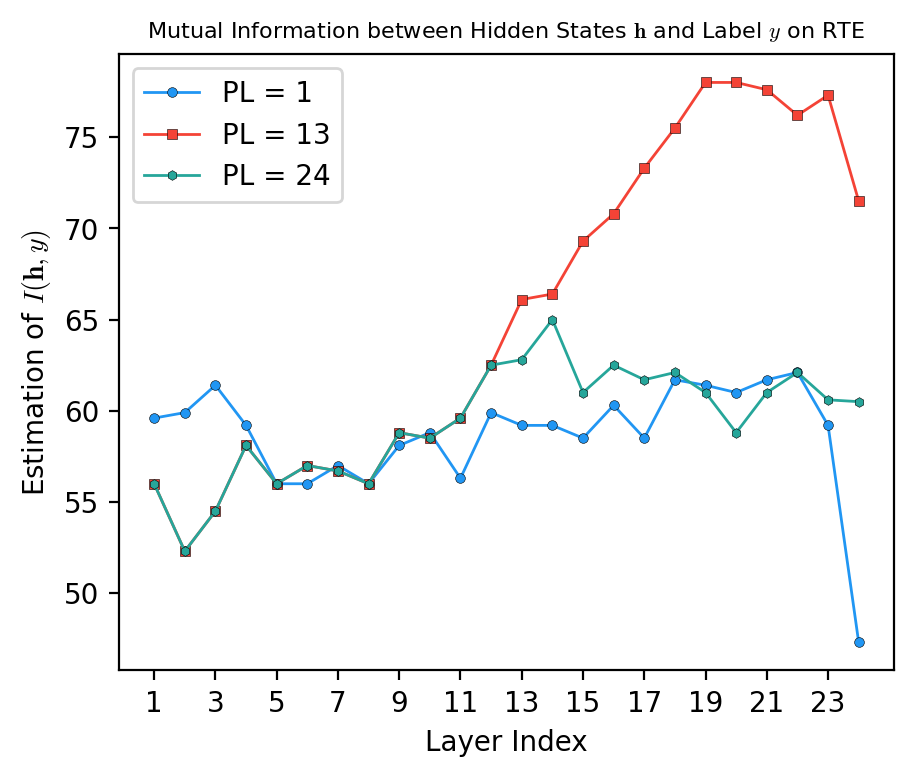}
    \end{subfigure}
    \caption{\textbf{Left}: The performance achieved by inserting a soft prompt into different layers of RoBERTa\textsubscript{LARGE}. \textbf{Middle}: Comparison of convergence rates for different prompt layers. \textbf{Right}: The estimated mutual information between hidden states of each layer and label. 'PL' denotes the prompt layer. 'PL = 1' denotes the traditional prompt tuning~\citep{lester2021pt}. We show mean and standard deviation of performance over 3 different random seeds.}
    \label{fig:late_prompt}
\end{figure*}

\paragraph{Does shortening the propagation distance improve performance?}
\label{para:late_prompt}

We start by considering a simple experiment setting where the soft prompt is inserted into different layers of RoBERTa\textsubscript{LARGE} then we look at how performance changes as the prompt layer changes. As shown in the left plots of Figure~\ref{fig:late_prompt}, we can observe that the performance first increases and then decreases with the rise of the prompt layer and obtain the highest performance when the prompting layer is in the range of 12 to 14. In addition, we also explore the convergence rates at different prompt layers. For simplification, we only consider three different prompt layers 1, 13, and 24. The middle plots in Figure~\ref{fig:late_prompt} show that the model has the fastest convergence rate when the prompt layer is 13. The trend is consistent with the performance trend shown on the left plots. We can preliminarily identify that properly shortening the propagation distance can improve performance according to these results. However, the performance starts to degrade when we extremely shorten the propagation path of task-related information. We attribute this to the interaction between the prompt and inputs becomes very weak when we unduly shorten the propagation path, which leads to the slighter influence of the prompt on model outputs and the gradual decline of performance.   

\paragraph{Task-related information in hidden states.}
To quantify the task-related information carried in the soft prompt, we follow \citet{wang2021lsl} and adopt the mutual information $I(\mathbf{h}, y)$ between the hidden states and label of each input. The estimate method of $I(\mathbf{h}, y)$ is provided in Appendix~\ref{sec:mutual_information}. The right plots of Figure~\ref{fig:late_prompt} show the $I(\mathbf{h}, y)$ at different layers. We note that $I(\mathbf{h}, y)$ gradually increases with the forward pass of prompt (i.e., the effect of the prompt on the hidden states gradually increases) when the prompt layer is 13. And its $I(\mathbf{h}, y)$ in the last layer is the highest among the three different prompt layer settings, which means that the soft prompt carries more task-related information. The other two prompt layer settings all collapse, especially on the RTE task, because there is no better trade-off between the propagation distance and the effect of prompt on hidden states.

The above observations suggest that our conjecture about the poor performance of prompt tuning is correct. The long propagation path of task-related information leads to poor performance and low convergence rate. And we find that properly shortening the propagation distance can improve performance.

\section{LPT: Late Prompt Tuning}
\label{sec:lip}

\begin{figure}[t!]
    \centering
    \includegraphics[width=\linewidth]{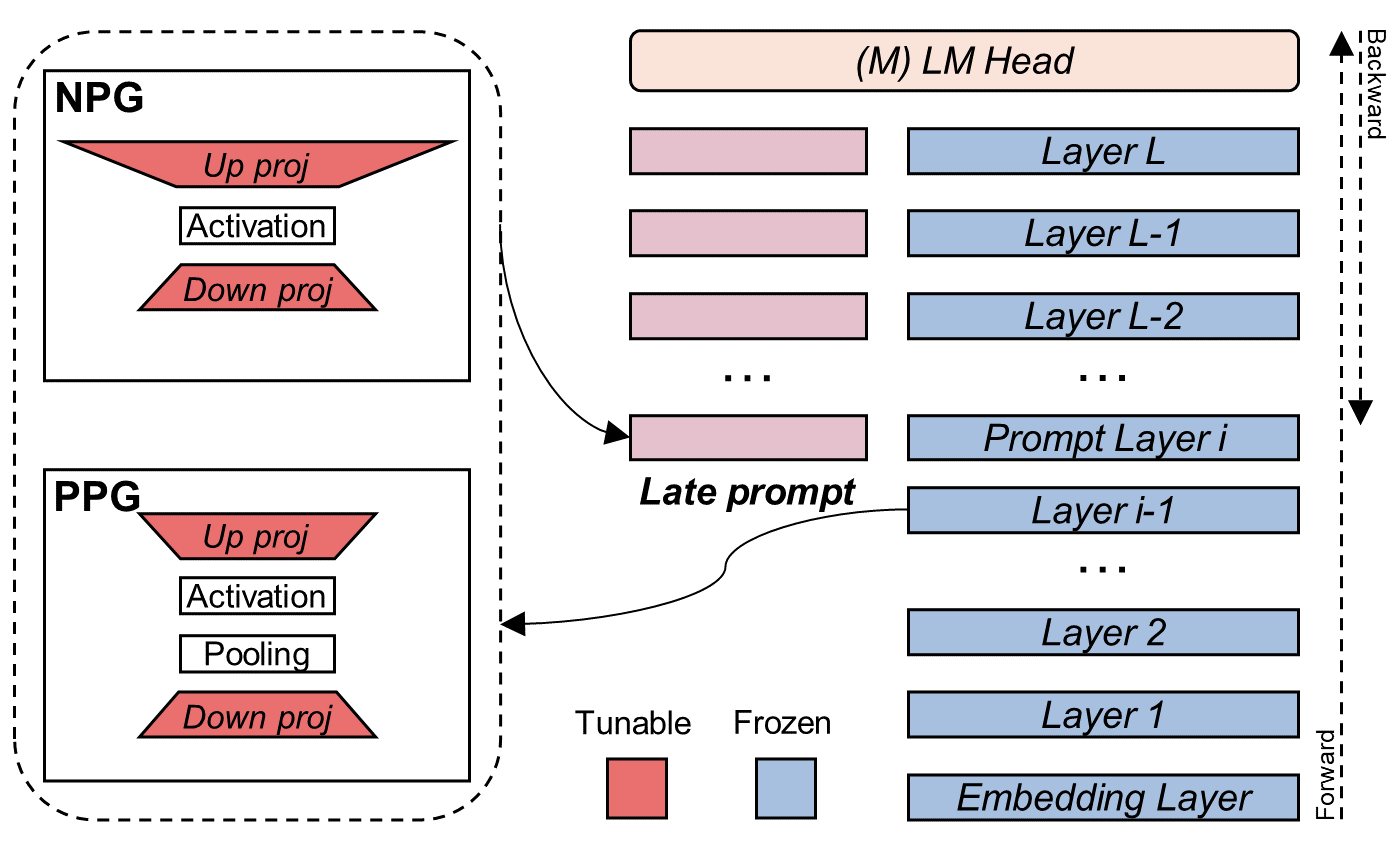}
    \caption{An illustration of LPT. \textbf{Left}: Naive (NPG) and pooling (PPG) prompt generators. \textbf{Right}: The forward and backward pass of LPT.}
    \label{fig:lip}
\end{figure}

From the experiment results in Section~\ref{sec:late_prompt}, we observe that using late prompt can greatly improve the performance of prompt tuning. Moreover, late prompt can bring two other advantages: \textbf{(1)} No gradient calculation for model parameters below the prompt layer; \textbf{(2)} The hidden states produced by the model before the prompt layer can be used to generate a great independent prompt for each instance. Based on these advantages, we propose an efficient prompt-based tuning method LPT which combines late and instance-aware prompts. An illustration of LPT is shown in Figure~\ref{fig:lip}. In this section, we will introduce two different prompt generators used in LPT and how to determine the prompt layer.

\subsection{Prompt Generators}

\paragraph{Naive prompt generator (NPG).} The prompt generator is a simple feed-forward layer with bottleneck architecture. Assume the prompt length is $l$, then we can generate an independent prompt for each instance as below:
\begin{align}
     \mathbf{\hat{p}} &= \mathbf{W_2}(\rm ReLU(\mathbf{W_1h}_{\texttt{[CLS]}} + \mathbf{b_1})) + \mathbf{b_2}, \\
     \mathbf{p} &= \rm Reshape(\mathbf{\hat{p}}),     
\end{align}
where $\mathbf{W_1} \in \mathbb{R}^{m \times d}$, $\mathbf{W_2} \in \mathbb{R}^{(l \times d) \times m}$, $\mathbf{h}_{\texttt{[CLS]}}$ $\in$ $\mathbb{R}^d$ and $\mathbf{p} \in \mathbb{R}^{l \times d}$. $\mathbf{b_1}$ and $\mathbf{b_2}$ are bias terms. $d$ is the dimension of hidden states. Since $m \ll d$, the prompt generator doesn't have too many parameters. However, the number of parameters within $\mathbf{W_2}$ will increase with the prompt length $l$. To tackle this problem, we propose the following pooling prompt generator. 

\paragraph{Pooling prompt generator (PPG).} PPG introduces a pooling operation between down-projection and up-projection operations, which directly obtains the prompt with length $l$ through pooling on input sequences (i.e., pooling the input with length $n$ to the prompt with length $l$). The generator is more lightweight to generate a prompt, 
\begin{align}
     \mathbf{\hat{p}} &= \rm ReLU(Pooling(\mathbf{W_1h} + \mathbf{b_1})), \\
     \mathbf{p} &= \mathbf{W_2}\mathbf{\hat{p}} + \mathbf{b_2}. 
\end{align}
Different from NPG, $\mathbf{W_1} \in \mathbb{R}^{m \times d}$, $\mathbf{W_2} \in \mathbb{R}^{d \times m}$ and $\mathbf{h} \in \mathbb{R}^{d \times n}$ here. $n$ is the length of the original input. In this paper, we consider both Average Pooling and Max Pooling, referred to as  \textbf{APPG} and \textbf{MPPG}, respectively.

\begin{figure}[t!]
    \centering
    \begin{subfigure}{0.49\linewidth}
    \centering
    \includegraphics[width=\linewidth]{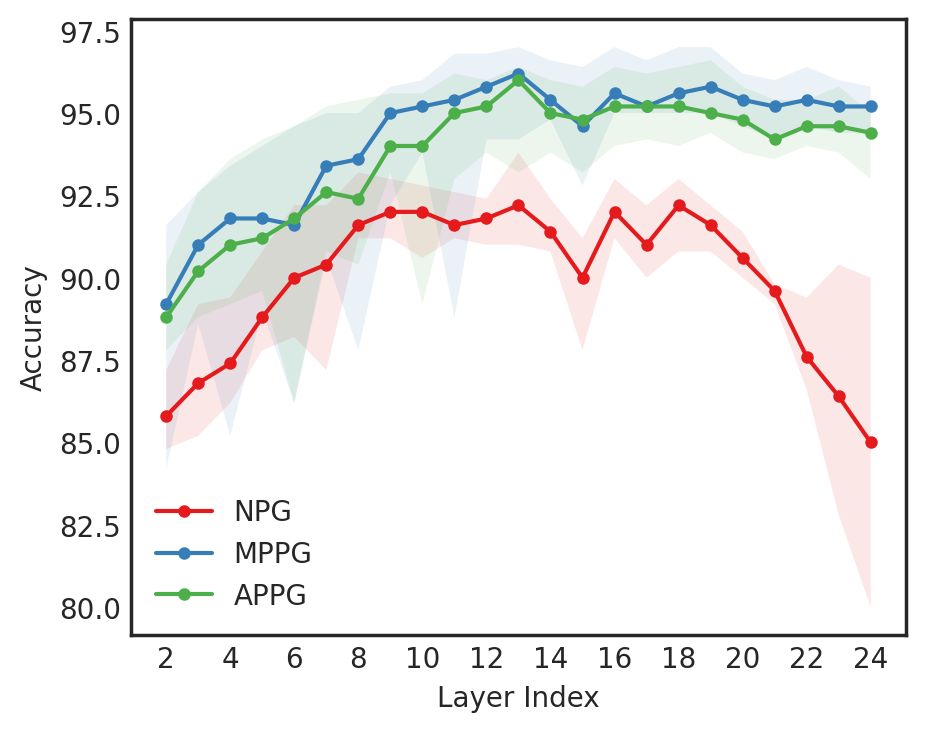}
    \caption{TREC}
    \end{subfigure}
    \begin{subfigure}{0.49\linewidth}
    \centering
    \includegraphics[width=\linewidth]{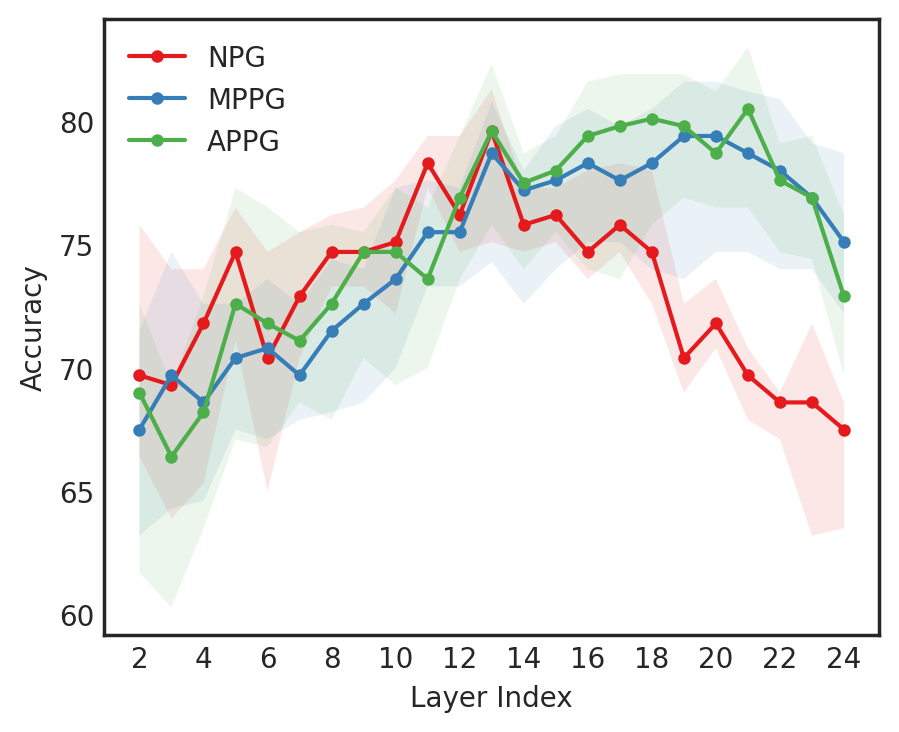}
    \caption{RTE}
    \end{subfigure}
    \caption{The change trend of performance with different prompt layers for three different prompt generators. The backbone model is RoBERTa\textsubscript{LARGE}. We show mean and standard deviation of performance over 3 different random seeds.}
    \label{fig:prompt_layer_roberta}
\end{figure}

\begin{table*}[t!]
\resizebox{\linewidth}{!}{
\begin{tabular}{ccccccccccccc}
\toprule
\multirow{2}{*}{\textbf{Method}} & \textbf{Tunable}    & \textbf{SST-2}      & \textbf{MPQA}       & \textbf{MR}         & \textbf{Subj}       & \textbf{TREC}       & \textbf{MNLI}       & \textbf{MRPC}         & \textbf{QNLI}       & \textbf{QQP}          & \textbf{RTE}        & \multirow{2}{*}{\textbf{Avg}} \\
                        & \textbf{Parameters} & (acc)      & (acc)      & (acc)      & (acc)      & (acc)      & (acc)      & (acc and F1) & (acc)      & (acc and F1) & (acc)      &                      \\
\midrule
Model Tuning            & \cellcolor[gray]{0.55}355M       & 95.6       & \textbf{90.2}       & 91.3       & 96.8      & \textbf{97.6}       & 89.3       & \textbf{91.2}         & 94.6       & \textbf{90.7}         & \textbf{86.2}       & \textbf{92.4}                 \\
\hdashline
Adapter                 & \cellcolor[gray]{0.65}1.6M       & \textbf{96.2} (0.2) & 89.2 (0.5) & 91.6 (0.4) & 96.8 (0.4) & 97.0 (0.3) & 9\textbf{0.5} (0.1) & 90.3 (1.0)   & \underline{94.7} (0.3) & 89.4 (0.7)   & \underline{85.5} (1.2) & \underline{92.3}                 \\
AdapterDrop             & \cellcolor[gray]{0.75}811K       & 95.3 (0.3) & 89.1 (0.7) & 91.0 (0.5) & 95.3 (0.6) & 95.7 (0.5) & 88.5 (0.2) & 90.1 (1.3)   & 93.3 (0.3) & 88.3 (0.3)   & 81.1 (2.0) & 90.8                 \\
\hdashline
BitFit             & \cellcolor[gray]{0.85}273K       & 95.9 (0.1) & 89.2 (0.9) & \underline{91.8} (0.5) & \underline{96.9} (0.1) & 96.2 (0.3) & \underline{90.0} (0.1) & 89.6 (0.9)   & 94.4 (0.2) & 87.9 (0.4)   & 82.4 (1.1) & 91.4                 \\
LoRA             & \cellcolor[gray]{0.75}788K       & \underline{96.2} (0.3) & 90.1 (0.3) & \textbf{92.0} (0.1) & \textbf{97.1} (0.4) & 96.8 (0.6) & 89.8 (0.3) & \underline{91.1} (0.6)   & \textbf{94.8} (0.2) & \underline{89.8} (0.1)   & 84.8 (2.1) & 92.3                 \\
\hdashline
Prompt Tuning           & \cellcolor[gray]{0.95}21K        & 94.9 (0.5) & 88.8 (0.8) & 89.6 (0.5) & 93.9 (0.6) & 86.4 (0.7) & 86.7 (0.9) & 75.7 (0.7)   & 91.4 (0.1) & 81.2 (0.8)   & 60.8 (0.5) & 84.9                 \\
Prompt Tuning-256       & \cellcolor[gray]{0.85}262K       & 95.8 (0.4) & \underline{90.2} (0.2) & 91.8 (0.4) & 95.8 (0.5) & 93.3 (0.4) & 87.7 (0.5) & 76.2 (2.4)   & 91.6 (0.8) & 85.3 (0.3)   & 59.7 (2.4) & 86.7                 \\
P-tuning v2             & \cellcolor[gray]{0.7}985K       & 95.8 (0.4) & 89.9 (0.6) & 91.4 (0.4) & 96.5 (0.2) & 95.8 (0.6) & 88.2 (0.2) & 86.5 (2.1)   & 93.7 (0.3) & 85.3 (0.2)   & 66.9 (2.3) & 89.0                 \\
S-IDPG-PHM              & \cellcolor[gray]{0.9}114K       & 94.8 (0.3) & 89.5 (0.6) & 90.8 (0.5) & 95.9 (0.6) & 89.3 (0.4) & 87.4 (0.5) & 77.3 (1.2)   & 91.2 (0.4) & 82.3 (1.9)   & 62.7 (1.9) & 86.1                 \\
\midrule
\multicolumn{12}{c}{\textit{LPT}} \\
\midrule
LPT w/o PG              & \cellcolor[gray]{0.95}21K        & 95.5 (0.3) & 87.6 (1.7) & 89.3 (0.6) & 95.1 (0.2) & 89.7 (0.7) & 88.0 (0.4) & 82.3 (1.3)   & 92.0 (0.1) & 84.2 (0.5)   & 75.2 (1.8) & 87.9                 \\
LPT w/ NPG            & \cellcolor[gray]{0.75}792K       & 95.5 (0.4) & 89.0 (0.1) & 90.9 (0.2) & 95.8 (0.2) & 95.9 (0.4) & 87.0 (0.3) & 88.4 (1.5)   & 91.7 (0.6) & 86.6 (0.5)   & 79.7 (3.2) & 90.1                 \\
LPT w/ MPPG           & \cellcolor[gray]{0.85}263K       & 95.4 (0.4) & 89.1 (0.2) & 90.7 (0.1) & 96.5 (0.2) & \underline{97.4} (0.2) & 87.7 (0.3) & 90.4 (0.6)   & 91.3 (0.3) & 88.6 (0.4)   & 78.7 (3.3) & 90.6                 \\
LPT w/ APPG           & \cellcolor[gray]{0.85}263K       & 95.3 (0.2) & 89.1 (0.3) & 90.7 (0.1) & 96.2 (0.2) & 97.0 (0.2) & 87.4 (0.3) & 90.2 (1.0)   & 91.6 (0.4) & 87.4 (0.4)   & 79.2 (3.3) & 90.4  \\  
\bottomrule
\end{tabular}
}
\caption{Overall comparison in full-data scenario. All the methods are evaluated on test sets except the tasks from GLUE benchmark. We report mean and standard deviation of performance over 3 different random seeds for all the methods except model tuning. The best results are highlighted in \textbf{bold} and the second best results are marked with \underline{underline}. Prompt Tuning-256 indicates the prompt tuning method with prompt length 256. All the results are obtained using RoBERTa\textsubscript{LARGE}.}
\label{tab:full_results}
\end{table*}

\subsection{How to Determine Prompt Layer?}
\label{subsec:prompt_layer}
Generating a good prompt needs a good contextual representation for the input. In this sub-section, we will explore how to choose the prompt layer to guarantee that LPT can attain a good trade-off between performance and efficiency through some pilot experiments on TREC~\citep{voorhees2000trec} and RTE~\citep{dagan2005rte} datasets. As shown in Figure~\ref{fig:prompt_layer_roberta}, the performance of NPG has a significant decline when the prompt layer is in the range from 14 to 24. However, different from NPG, APPG and MPPG retain high performance as the prompt layer approaches the output layer, especially on TREC dataset. We believe that this is due to the hidden states from the higher layers can help generate a better prompt, while NPG only uses \texttt{[CLS]} token as the representation of the entire input when generating the prompt, which leads to the loss of information. According to the above observations, LPT with APPG and MPPG can achieve a better trade-off for both relatively simple (TREC) and difficult (RTE) tasks. But in this work,  to ensure that all methods (NPG, APPG and MPPG) can achieve a good performance while maintaining a relatively low training costs, we simply choose the most intermediate layer of PTM as the prompt layer. That is, we choose the 13-th layer as the prompt layer for RoBERTa\textsubscript{LARGE}. 

\section{Experiments}

\subsection{Evaluation Datasets}
We evaluate our method on 5 single-sentence and 5 sentence-pair classification tasks, including 6 tasks from GLUE benchmark~\citep{wang2019glue} and 4 other popular tasks include MPQA~\citep{wiebe2005mpqa}, MR~\citep{pang2005mr}, Subj~\citep{pang2004subj} and TREC~\citep{voorhees2000trec} tasks. All details about data statistics and splits can be found in Appendix~\ref{sec:dataset_appendix}. 

\subsection{Experiment Settings}
We evaluate our method in both full-data and few-shot scenarios on three PTMs, including RoBERTa\textsubscript{LARGE}~\citep{liu2019roberta}, DeBERTa\textsubscript{LARGE}~\citep{he2021deberta} and GPT2\textsubscript{LARGE}~\citep{radford2019gpt2}. According to the conclusion from the Section~\ref{subsec:prompt_layer}, we choose the 13-th layer as the prompt layer for RoBERTa\textsubscript{LARGE} and DeBERTa\textsubscript{LARGE}, and the 19-th layer for GPT2\textsubscript{LARGE} except special explanation. More implementation details are provided in Appendix~\ref{sec:hyper&template}.

\subsection{Baselines}
We consider \textit{Model Tuning}, \textit{adapter-based tuning}, \textit{prompt-based tuning} methods and two other state-of-the-art PETuning methods that include \textbf{(1) BitFit}~\citep{Zaken2022bitfit} and \textbf{(2) LoRA}~\citep{hu2022lora} as our baselines. For adapter-based tuning methods, we compare with  \textbf {(1) Adapter}~\citep{Houlsby2019adapter} and \textbf{(2) AdapterDrop}~\citep{Andreas2020adapterdrop}. For prompt-based tuning methods, we compare with \textbf{(1) Prompt Tuning}~\citep{lester2021pt}, \textbf{(2) P-tuning v2}~\citep{Liu2022ptuningv2} and \textbf{(3) IDPG}~\citep{Wu2022idpg}. We implement Aadpter, AdapterDrop, BitFit,  and LoRA using OpenDelta\footnote{\url{https://github.com/thunlp/OpenDelta}} library. For IDPG which also raises instance-aware prompt, we only compare with the version with single-layer prompt, that is S-IDPG-PHM. And we don't use supplementary training like \citet{Wu2022idpg} to enhance performance.

\begin{table*}
\resizebox{\linewidth}{!}{
\begin{tabular}{ccccccccccccc}
\toprule
\multirow{2}{*}{\textbf{Method}} & \textbf{Tunable}    & \textbf{SST-2}      & \textbf{MPQA}       & \textbf{MR}         & \textbf{Subj}       & \textbf{TREC}       & \textbf{MNLI}       & \textbf{MRPC}         & \textbf{QNLI}       & \textbf{QQP}          & \textbf{RTE}        & \multirow{2}{*}{\textbf{Avg}} \\
                        & \textbf{Parameters} & (acc)      & (acc)      & (acc)      & (acc)      & (acc)      & (acc)      & (acc and F1) & (acc)      & (acc and F1) & (acc)      &                      \\
\midrule
Model Tuning            & \cellcolor[gray]{0.55}355M       & 89.6 (1.2) & 81.5 (2.0) & 85.5 (2.5) & \textbf{93.6} (0.5) & \underline{91.3} (1.9) & 51.5 (3.3) & 78.3 (1.0)   & \underline{73.9} (6.6) & 71.6 (2.9)   & 48.6 (3.0) & 76.5                 \\
\hdashline
Adapter                 & \cellcolor[gray]{0.65}1.6M       & 90.8 (1.3) & 81.8 (3.0) & 86.3 (1.5) & 93.4 (0.8) & 89.7 (3.7) & 42.9 (1.2) & 77.9 (2.4)   & 61.5 (2.7) & 67.3 (2.0)   & 52.0 (2.4) & 74.4                 \\
AdapterDrop             & \cellcolor[gray]{0.75}811K       & 87.8 (1.2) & 81.4 (2.3) & 85.8 (1.5) & \underline{93.5} (0.9) & 89.8 (4.6) & 41.0 (0.8) & 76.6 (0.9)   & 60.4 (3.9) & 64.7 (2.5)   & 50.4 (1.8) & 73.1                 \\
\hdashline
BitFit             & \cellcolor[gray]{0.85}273K       & \underline{91.8} (0.9) & 84.0 (2.1) & 86.9 (1.0) & 92.3 (1.0) & 90.8 (1.8) & 42.0 (0.9) & 77.0 (2.7)   & 60.3 (6.5) & 64.9 (0.9)   & 50.8 (2.2) & 74.1                 \\
LoRA             & \cellcolor[gray]{0.75}788K       & 91.0 (1.3) & 83.2 (1.3) & 87.4 (0.7) & 92.6 (1.4) & \textbf{92.0} (0.4) & 48.1 (3.7) & \underline{78.5} (1.7)   & 65.9 (5.7) & 69.7 (3.3)   & 51.0 (2.0) & 75.9                 \\
\hdashline
Prompt Tuning           & \cellcolor[gray]{0.95}21K        & 90.0 (2.2) & 73.5 (5.8) & 85.1 (1.2) & 80.6 (3.8) & 72.3 (4.7) & 47.3 (1.8) & 74.2 (1.0)   & 55.8 (1.7) & 52.7 (2.1)   & 59.6 (2.3) & 69.1                 \\
P-tuning v2             & \cellcolor[gray]{0.7}985K       & 89.4 (0.6) & 80.6 (1.6) & 84.6 (2.3) & 91.7 (1.4) & 84.9 (4.5) & 37.3 (1.8) & 75.0 (0.6)   & 54.2 (1.1) & 60.7 (3.0)   & 54.9 (2.1) & 71.3                 \\
S-IDPG-PHM              & \cellcolor[gray]{0.9}114K       & 90.5 (1.7) & 75.5 (5.8) & 85.8 (0.8) & 81.6 (1.8) & 75.3 (3.7) & 47.8 (1.6) & 75.2 (1.1)   & 56.9 (0.9) & 54.5 (1.9)   & 59.8 (2.4) & 70.3                 \\
\midrule
\multicolumn{12}{c}{\textit{LPT}} \\
\midrule
LPT w/o PG              & \cellcolor[gray]{0.95}21K        & 91.3 (1.0) & 80.6 (7.3) & 88.2 (0.5) & 90.7 (0.9) & 79.9 (1.5) & 52.9 (5.5) & 77.4 (1.0)   & 66.2 (3.0) & 68.2 (4.0)   & 58.3 (2.9) & 75.4                 \\
LPT w/ NPG            & \cellcolor[gray]{0.75}792K       & \textbf{92.7} (0.8) & \textbf{86.8} (1.4) & \textbf{88.5} (0.5) & 92.7 (0.5) & 90.9 (2.5) & \textbf{64.3} (2.0) & \textbf{80.6} (2.0)   & \textbf{75.7} (3.2) & \textbf{74.6} (1.9)   & \textbf{68.1} (5.5) & \textbf{81.5}                 \\
LPT w/ MPPG           & \cellcolor[gray]{0.85}263K       & 90.2 (0.9) & 83.9 (5.0) & \underline{88.5} (0.6) & 92.7 (0.9) & 85.9 (5.3) & 58.8 (1.6) & 77.3 (1.5)   & 71.9 (2.9) & \underline{72.8} (2.3)   & 63.0 (3.4) & 78.5                 \\
LPT w/ APPG           & \cellcolor[gray]{0.85}263K       & 90.4 (0.7) & \underline{84.4} (6.1) & 88.3 (0.6) & 92.6 (1.2) & 87.9 (3.7) & \underline{60.1} (2.4) & 78.2 (2.6)   & 71.6 (4.1) & 72.0 (2.0)   & \underline{64.0} (2.9) & \underline{79.0} \\
\bottomrule
\end{tabular}
}
\caption{Results in the few-shot scenario of 100 training samples. We report mean and standard deviation of performance over 4 different data splits for all the methods. \textbf{Bold} and \underline{Underline} indicate the best and the second best results. All the results are obtained using RoBERTa\textsubscript{LARGE}.}
\label{tab:100_results}
\end{table*}

\begin{table*}[t]
\resizebox{\linewidth}{!}{
\begin{tabular}{ccccccccccccc}
\toprule
\multirow{2}{*}{\textbf{Method}} & \textbf{Tunable}    & \textbf{SST-2}      & \textbf{MPQA}       & \textbf{MR}         & \textbf{Subj}       & \textbf{TREC}       & \textbf{MNLI}       & \textbf{MRPC}         & \textbf{QNLI}       & \textbf{QQP}          & \textbf{RTE}        & \multirow{2}{*}{\textbf{Avg}} \\
                        & \textbf{Parameters} & (acc)      & (acc)      & (acc)      & (acc)      & (acc)      & (acc)      & (acc and F1) & (acc)      & (acc and F1) & (acc)      &                      \\
\midrule
Model Tuning            & \cellcolor[gray]{0.55}355M       & 91.4 (0.8) & 87.2 (1.1) & \underline{89.4} (0.6) & \textbf{95.1} (0.4) & 95.4 (0.5) & \underline{75.3} (2.1) & \textbf{85.1} (1.8)   & \textbf{85.2} (0.9) & \underline{77.3} (1.2)   & 67.0 (7.7) & \underline{84.8}                 \\
\hdashline
Adapter                 & \cellcolor[gray]{0.65}1.6M       & 92.0 (1.0) & 86.5 (1.5) & 88.4 (1.0) & \underline{95.1} (0.4) & 95.0 (0.4) & 70.5 (4.8) & 83.6 (1.1)   & 78.0 (1.8) & 72.1 (6.7)   & 67.5 (6.7) & 82.9                 \\
AdapterDrop             & \cellcolor[gray]{0.75}811K       & 91.2 (1.0) & 84.4 (1.2) & 88.4 (0.8) & 95.1 (0.4) & \textbf{95.7} (0.4) & 66.1 (4.5) & 82.5 (1.6)   & 78.9 (1.0) & 73.4 (0.6)   & 62.0 (3.2) & 81.6                 \\
\hdashline
BitFiT                 & \cellcolor[gray]{0.85}273K       & \underline{92.2} (1.0) & \underline{87.6} (0.9) & 89.0 (0.8) & 94.7 (0.2) & 95.0 (0.6) & 73.0 (2.8) & 83.5 (0.6)   & 80.4 (1.4) & 75.6 (1.4)   & 59.0 (1.8) & 82.7                 \\
LoRA             & \cellcolor[gray]{0.75}788K       & 92.1 (1.1) & 87.5 (0.7) & 88.6 (1.4) & 95.1 (0.2) & \underline{95.5} (0.9) & 74.5 (2.9) & \underline{84.1} (0.6)   & 82.5 (1.5) & 76.4 (1.1)   & 62.8 (3.2) & 83.9 \\ 
\hdashline
Prompt Tuning           & \cellcolor[gray]{0.95}21K        & 91.1 (1.5) & 74.7 (5.1) & 88.3 (0.6) & 86.4 (0.4) & 81.7 (2.4) & 45.5 (1.5) & 74.6 (0.3)   & 58.1 (1.6) & 52.6 (5.8)   & 61.2 (1.7) & 71.4                 \\
P-tuning v2             & \cellcolor[gray]{0.7}985K       & 91.3 (0.3) & 85.1 (1.6) & 88.0 (1.5) & 94.5 (0.4) & 94.6 (0.8) & 61.6 (2.7) & 76.6 (1.8)   & 73.7 (2.4) & 71.7 (1.8)   & 56.0 (1.1) & 79.3                 \\
S-IDPG-PHM              & \cellcolor[gray]{0.9}114K       & 91.3 (0.5) & 75.9 (3.8) & 88.7 (0.4) & 87.2 (0.6) & 84.7 (2.1) & 46.3 (1.1) & 75.1 (0.8)   & 59.4 (0.7) & 56.4 (3.0)   & 64.7 (1.7) & 73.0                 \\
\midrule
\multicolumn{12}{c}{\textit{LPT}} \\
\midrule
LPT w/o PG              & \cellcolor[gray]{0.95}21K        & 91.9 (0.4) & 83.6 (1.0) & 88.7 (0.6) & 92.5 (0.7) & 84.2 (0.8) & 54.5 (5.8) & 80.0 (0.8)   & 75.3 (2.2) & 73.1 (1.9)   & 64.8 (3.1) & 78.9                 \\
LPT w/ NPG            & \cellcolor[gray]{0.75}792K       & \textbf{92.6} (0.4) & \textbf{87.8} (0.5) & \textbf{90.0} (0.4) & 94.9 (0.2) & 93.5 (0.4) & \textbf{76.0} (1.0) & 81.4 (0.9)   & \underline{83.2} (1.3) & \textbf{77.9} (0.8)   & \textbf{74.7} (2.7) & \textbf{85.2}                 \\
LPT w/ MPPG           & \cellcolor[gray]{0.85}263K       & 91.0 (0.8) & 86.3 (1.0) & 89.3 (0.3) & 94.6 (0.3) & 93.2 (0.9) & 70.9 (3.5) & 82.5 (0.6)   & 78.1 (2.0) & 75.1 (1.1)   & 69.0 (3.3) & 83.0                 \\
LPT w/ APPG           & \cellcolor[gray]{0.85}263K       & 91.9 (0.3) & 86.2 (0.9) & 89.0 (0.3) & 94.3 (0.2) & 92.5 (1.2) & 69.2 (3.5) & 82.2 (1.3)   & 79.4 (1.9) & 74.8 (1.3)   & \underline{70.4} (1.4) & 83.0 \\
\bottomrule
\end{tabular}
}
\caption{Results in the few-shot scenario of 500 training samples. We report mean and standard deviation of performance over 4 different data splits for all the methods. \textbf{Bold} and \underline{Underline} indicate the best and the second best results. All the results are obtained using RoBERTa\textsubscript{LARGE}.}
\label{tab:500_results}
\end{table*}

\subsection{Main Results}
\paragraph{Results in full-data scenario.} The overall comparison of the results in full-data scenario is shown in Table~\ref{tab:full_results}. We can observe that: (\romannumeral1) Our method with only late prompt, that is LPT w/o PG can greatly improve the performance of the traditional prompt tuning under the same number of tunable parameters and even is comparable with P-tuning v2 which inserts prompts to each layer of PTM. (\romannumeral2) Increasing the prompt length for prompt tuning can improve performance to some extend, but increasing the training burden and inference latency notably. (\romannumeral3) Our method LPT with different prompt generators (i.e., LPT w/ NPG, LPT w/ MPPG, and LPT w/ APPG) outperforms all the prompt-based methods including S-IDPG-PHM that also claims instance-aware prompt. (\romannumeral4) The performance of LPT with the prompt generators is comparable with AdapterDrop, especially for LPT w/ MPPG and LPT w/ APPG. But their number of tunable parameters is only one-third of AdapterDrop. (\romannumeral5) Prompt-based methods are weaker than adapter-based methods and model tuning on sentence-pair tasks, which is consistent with the results from \citet{Sun2022bbt} and \citet{Ding2022deltatuning}. It may be because sentence-pair tasks are more difficult than single-sentence tasks and more influenced by manual templates and label words.

\paragraph{Results in few-shot scenario} We further evaluate our method in few-shot scenario. Following \citet{Wu2022idpg}, we consider two settings where the number of training data is 100 and 500, respectively. We randomly sample the training samples from original training sets. Besides, we randomly sample 1000 samples from the original training sets as development sets and there is no overlap with sampled training sets. For the tasks from GLUE benchmark~\citep{wang2019glue}, the original development sets are used as the test sets and the test sets remain unchanged for 4 other tasks.

Table~\ref{tab:100_results} and~\ref{tab:500_results} show the overall comparison of all the methods in the few-shot scenario. LPT w/ NPG outperforms all the baselines in two different few-shot settings. Especially when the training set has only 100 samples, LPT w/ NPG outperforms model tuning by 5 points and Adapter by 7.1 points. This indicates that our method has better generalization performance when the training data is very scarce. However, we note that LPT w/ MPPG and LPT w/ APPG don't perform as well in the few-shot scenario as they do in the full-data scenario. We speculate that this is owing to the optimal state of the pooling layer to retain only useful information, and sufficient training data is needed to achieve this state. Nevertheless, both LPT w/ MPPG and LPT w/ APPG are also superior to all the baselines when the training set has 100 samples.

\paragraph{Results on other PTMs} To verify the generality of our conclusion about why prompt tuning performs poorly and the versatility of the proposed method LPT, we also conduct experiments on two other popular PTMs, DeBERTa\textsubscript{LARGE}~\citep{he2021deberta}, and GPT2\textsubscript{LARGE}~\citep{radford2019gpt2}. The results are shown in Table~\ref{tab:results_models}. Only using the late prompt to shorten the propagation path of task-related information (i.e., LPT w/o PG) is also far superior to the traditional prompt tuning method on these two PTMs. This result enhances the reliability of our conclusion. Moreover, LPT with different prompt generators further improves the performance, closing the gap with model tuning. 

\begin{table}[t]
\resizebox{\linewidth}{!}{
\begin{tabular}{ccccccc}
\toprule
\multirow{2}{*}{\textbf{Method}} & \textbf{Tunable}              & \textbf{Subj}       & \textbf{TREC}       & \textbf{MRPC}         & \textbf{RTE}        & \multirow{2}{*}{\textbf{Avg}} \\
                        & \textbf{Parameters}           & (acc)      & (acc)      & (acc and F1) & (acc)      &                      \\
\midrule
\multicolumn{7}{c}{\textit{DeBERTa\textsubscript{LARGE}}} \\  \midrule                                              
Model Tuning            & \cellcolor[gray]{0.6}406M & \textbf{97.4}       & \textbf{97.4}       & \textbf{91.2}         & \textbf{87.5}       & \textbf{93.4}                 \\
\hdashline
Prompt Tuning           & \cellcolor[gray]{0.95}21K & 94.2 (0.5) & 87.7 (2.0) & 79.8 (1.6)   & 64.6 (3.7) & 81.6                 \\
\hdashline
LPT w/o PG              & \cellcolor[gray]{0.95}21K & 94.9 (0.5) & 94.4 (0.3) & 81.4 (1.2)   & 75.1 (1.9) & 86.5                 \\
LPT w/ NPG            & \cellcolor[gray]{0.75} 792K                    & 96.5 (0.2) & 96.3 (0.3) & \underline{90.8} (0.8)   & \underline{84.4} (0.7) & \underline{92.0}                 \\
LPT w/ MPPG           & \cellcolor[gray]{0.85} 263K                    & \underline{96.9} (0.2) & \underline{97.3} (0.3) & 89.6 (1.0)   & 81.1 (1.6) & 91.2                 \\
LPT w/ APPG           & \cellcolor[gray]{0.85} 263K                    & 96.5 (0.2) & 97.0 (0.2) & 89.7 (1.2)   & 82.6 (1.3) & 91.5                 \\
\midrule
\multicolumn{7}{c}{\textit{GPT2\textsubscript{LARGE}}} \\
\midrule
Model Tuning            &  \cellcolor[gray]{0.6}  774M                  & \textbf{97.2}       & \textbf{97.0}         & \textbf{88.0}         & \textbf{75.8}       & \textbf{89.5}                 \\
\hdashline
Prompt Tuning           &  \cellcolor[gray]{0.95}  26K                  & 88.8 (1.0) & 82.7 (1.1) & 75.1 (0.5)   & 53.7 (1.3) & 75.1                 \\
\hdashline
LPT w/o PG              &  \cellcolor[gray]{0.95}  26K                  & 94.9 (1.2) & 93.7 (2.3) & 77.3 (1.3)   & 57.8 (2.1) & 80.9                 \\
LPT w/ NPG            &  \cellcolor[gray]{0.75}  990K                  & \underline{96.0} (0.3) & 96.1 (0.4) & 82.9 (1.0)   & 69.9 (1.0) & 86.2                 \\
LPT w/ MPPG           &  \cellcolor[gray]{0.85}  329K                  & 95.9 (0.3) & 96.3 (0.5) & 85.6 (0.4)   & 71.6 (0.6) & 87.4                 \\
LPT w/ APPG           &  \cellcolor[gray]{0.85}  329K                  & 95.6 (0.3) & \underline{96.7} (0.3) & \underline{85.7} (0.2)   & \underline{72.9} (0.8) & \underline{87.7} \\
\bottomrule
\end{tabular}
}
\caption{Results on two single-sentence and two sentence-pair tasks using DeBERTa\textsubscript{LARGE} and GPT2\textsubscript{LARGE} models as the backbone. \textbf{Bold} and \underline{Underline} indicate the best and the second best results.}
\label{tab:results_models}
\end{table}

\begin{table}[t]
\resizebox{\linewidth}{!}{
\begin{tabular}{cccccc}
\toprule
 \multirow{2}{*}{\textbf{Method}} & \multirow{2}{*}{\textbf{Accuracy}} & \textbf{Tuable}    & \textbf{Training Speed} & \textbf{Memory Cost} \\
                         &                           & \textbf{Parameters} & tokens/ms (\textcolor{blue}{$\uparrow$})     & GB (\textcolor{blue}{$\downarrow$})        \\
\midrule
\multicolumn{6}{c}{\textit{RoBERTa\textsubscript{LARGE}}} \\
\midrule
 Model Tuning            & 52.0 (1.9)                & \cellcolor[gray]{0.55}355M       & 11.6         & 23.5      \\
 \hdashline
Adapter                 & 50.3 (2.5)                & \cellcolor[gray]{0.65}1.6M       & 15.5 (\textcolor{blue}{1.3$\times$})    & 16.5 (\textcolor{blue}{29.8\%}) \\
AdapterDrop             & 49.4 (3.4)                & \cellcolor[gray]{0.75}811K       & 21.6 (\textcolor{blue}{1.9$\times$})    & \; 9.5    (\textcolor{blue}{59.6\%})  \\
\hdashline
BitFit             & 50.2 (1.8)                & \cellcolor[gray]{0.85}273K       & 16.5 (\textcolor{blue}{1.4$\times$})     & 15.7 (\textcolor{blue}{33.2\%})  \\
LoRA             & 50.1 (2.7)                & \cellcolor[gray]{0.75}788K       & 16.4 (\textcolor{blue}{1.4$\times$})     & 16.2 (\textcolor{blue}{31.1\%})  \\
\hdashline
Prompt Tuning           & 58.2 (1.7)                & \cellcolor[gray]{0.95}21K        & 16.9 (\textcolor{blue}{1.5$\times$})    & 17.8 (\textcolor{blue}{24.3\%}) \\
P-tuning v2             & 53.2 (2.4)                         & \cellcolor[gray]{0.7}985K       & 19.2 (\textcolor{blue}{1.7$\times$})              & 16.8 (\textcolor{blue}{28.5\%})         \\
S-IDPG-PHM              & 58.8 (1.9)                & \cellcolor[gray]{0.9}114K       & 12.0 (\textcolor{blue}{1.0$\times$})     & 16.8 (\textcolor{blue}{28.5\%}) \\
\hdashline
LPT w/ NPG            & \textbf{69.5} (3.1)                & \cellcolor[gray]{0.75}792K       & 23.2 (\textcolor{blue}{2.0$\times$})    & 10.1 (\textcolor{blue}{56.6\%}) \\
LPT w/ MPPG           & 62.4 (3.1)                & \cellcolor[gray]{0.85}263K       & 23.4 (\textcolor{blue}{2.0$\times$})    & 10.6 (\textcolor{blue}{54.9\%}) \\
LPT w/ APPG           & \underline{63.0} (2.2)                & \cellcolor[gray]{0.85}263K       & 23.4 (\textcolor{blue}{2.0$\times$})    & 10.6 (\textcolor{blue}{54.9\%}) \\
\midrule
\multicolumn{6}{c}{\textit{GPT2\textsubscript{LARGE}}} \\
\midrule
 Model Tuning            & 50.0 (1.9)                & \cellcolor[gray]{0.55}774M       & 2.6          & 22.1      \\
\hdashline
Adapter                 & 52.8 (2.9)                & \cellcolor[gray]{0.65}3.0M       & 3.3 (\textcolor{blue}{1.3$\times$})     & 11.8 (\textcolor{blue}{46.6\%}) \\
AdapterDrop             & 49.9 (0.9)                & \cellcolor[gray]{0.75}1.5M       & 6.0 (\textcolor{blue}{2.3$\times$})     & \; 8.4 (\textcolor{blue}{62.0\%})  \\
\hdashline
BitFit             & 51.3 (2.4)                & \cellcolor[gray]{0.83}511K       & 4.3 (\textcolor{blue}{1.7$\times$})     & 11.5 (\textcolor{blue}{48.0\%})  \\
LoRA             & 52.6 (1.9)                & \cellcolor[gray]{0.81}740K       & 4.1 (\textcolor{blue}{1.6$\times$})     & 11.5 (\textcolor{blue}{47.1\%})  \\
\hdashline
Prompt Tuning           & 50.3 (1.2)                & \cellcolor[gray]{0.95}26K        & 4.4 (\textcolor{blue}{1.7$\times$})     & 13.6 (\textcolor{blue}{38.5\%}) \\
P-tuning v2             & 49.7 (1.9)                         & \cellcolor[gray]{0.71}1.9M       & 4.5 (\textcolor{blue}{1.7$\times$})              & 13.0 (\textcolor{blue}{41.2\%})         \\
S-IDPG-PHM              & 52.1 (2.3)                & \cellcolor[gray]{0.9}171K       & 3.2 (\textcolor{blue}{1.2$\times$})     & 12.7 (\textcolor{blue}{42.5\%}) \\
\hdashline
LPT w/ NPG            & \textbf{56.9} (2.0)                & \cellcolor[gray]{0.79}990K       & 6.0 (\textcolor{blue}{2.3$\times$})     & \; 9.4 (\textcolor{blue}{57.5\%})  \\
LPT w/ MPPG           & \underline{54.2} (2.6)                & \cellcolor[gray]{0.85}329K       & 6.2 (\textcolor{blue}{2.4$\times$})     & \; 9.6 (\textcolor{blue}{56.6\%})  \\
LPT w/ APPG           & 53.6 (1.7)                & \cellcolor[gray]{0.85}329K       & 6.2 (\textcolor{blue}{2.4$\times$})     & \; 9.6 (\textcolor{blue}{56.6\%})  \\
\bottomrule
\end{tabular}
}
\caption{Comparison of parameter efficiency, training efficiency and memory cost for all the methods on two different backbone models. All methods are evaluated on RTE dataset.}
\label{tab:efficiency}
\end{table}

\subsection{Efficiency Evaluation}
\label{sec:efficiency}
We compare the efficiency of our method with all the baselines on RoBERTa\textsubscript{LARGE}~\citep{liu2019roberta} and GPT2\textsubscript{LARGE}~\citep{radford2019gpt2} models. For each backbone, we select the largest batch size such that model tuning method can fit the fixed budget of a NVIDIA GTX 3090 GPU (24GB) and other methods use the same batch size as model tuning. We set the length of all inputs to 256 and evaluate the accuracy in the few-shot scenario that the number of training data is 100 for all methods. 

In Table~\ref{tab:efficiency}, we report accuracy, tuable parameters, training speed (tokens per millisecond) and memory cost (GB) of each method. Our methods not only outperform all prompt-based methods considered in terms of efficiency and memory cost, but obtain the highest performance. Compared with AdapterDrop that has similar efficiency with LPT, our method LPT w/ NPG outperforms it by 20.1 and 7 points on RoBERTa\textsubscript{LARGE} and GPT2\textsubscript{LARGE}, respectively. In addition, we also explore the impact of the choice of prompt layer on all efficiency metrics, and the specific experiment results are in Appendix~\ref{sec:efficiency_pl}. Overall, given a large scale PTM with millions or billions of parameters, such as RoBERTa~\citep{liu2019roberta}, DeBERTa~\citep{he2021deberta}, and GPT2~\citep{radford2019gpt2}, higher training speed and lower memory cost is a paramount importance for practical applications. And LPT offers a better trade-off in terms of training budget and performance.

\subsection{Analyses}
\begin{figure}[t]
    \centering
    \begin{subfigure}{0.49\linewidth}
    \centering
    \includegraphics[width=\linewidth]{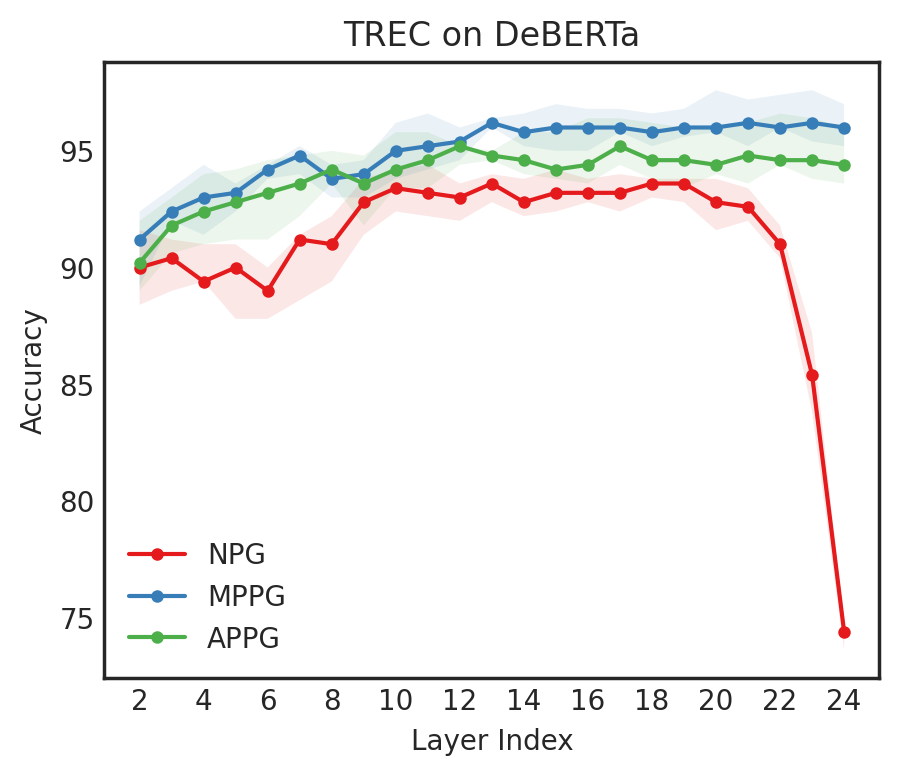}
    \end{subfigure}
    \begin{subfigure}{0.49\linewidth}
    \centering
    \includegraphics[width=\linewidth]{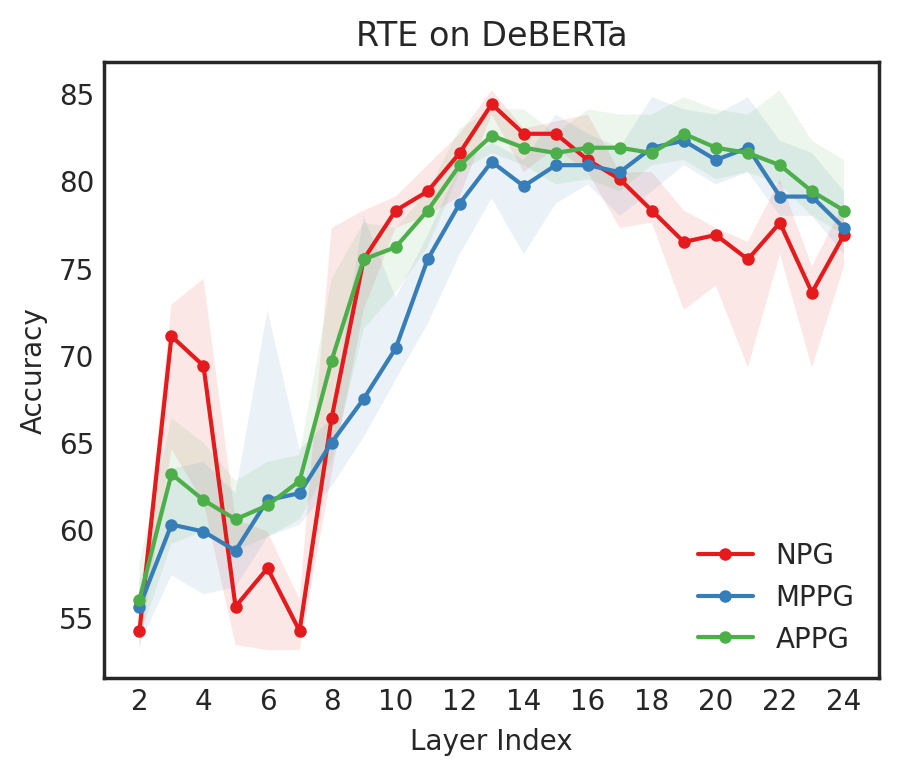}
    \end{subfigure}
    \\
    \centering
    \begin{subfigure}{0.49\linewidth}
    \centering
    \includegraphics[width=\linewidth]{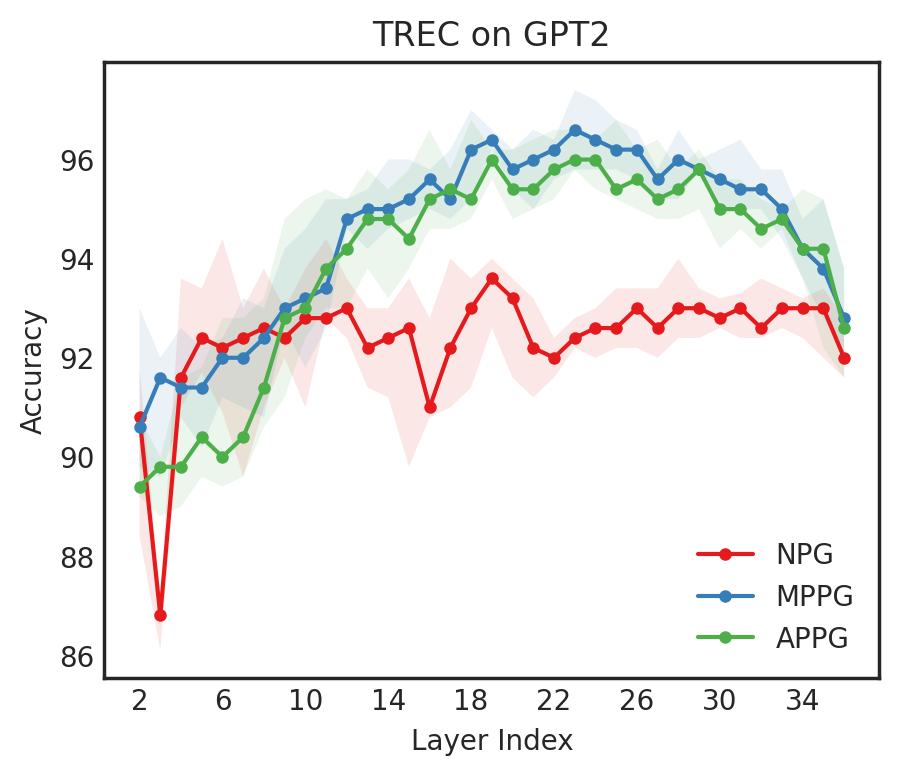}
    \end{subfigure}
    \begin{subfigure}{0.49\linewidth}
    \centering
    \includegraphics[width=\linewidth]{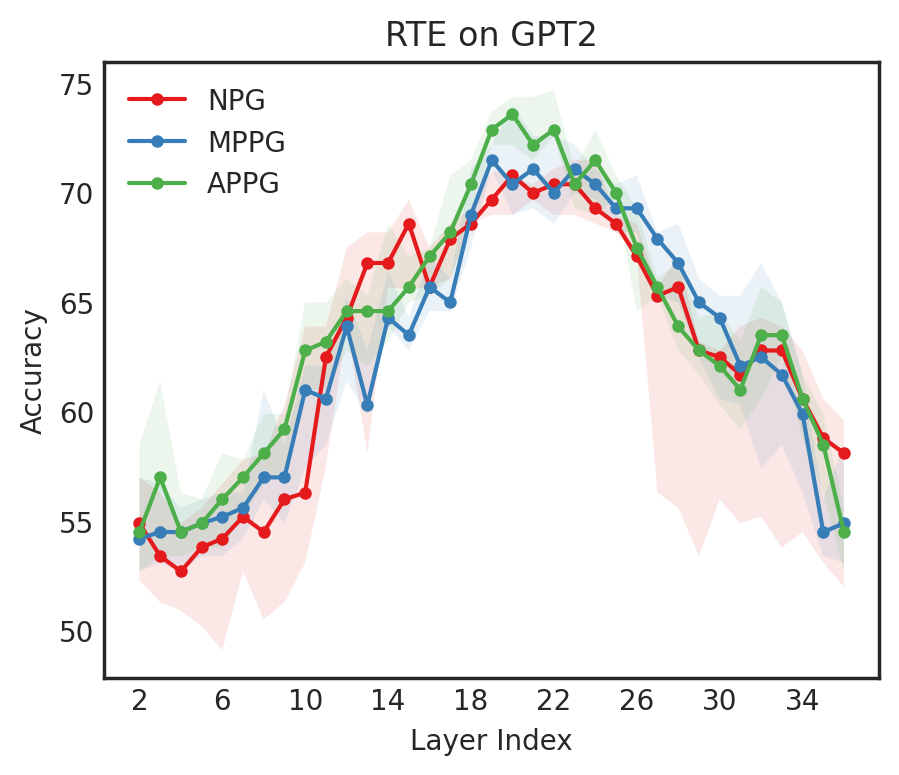}
    \end{subfigure}
    \caption{The change trend of performance with different prompt layers on DeBERTa\textsubscript{LARGE} (upper) and GPT2\textsubscript{LARGE} (lower). We show mean and standard deviation of performance over 3 different random seeds.}
\label{fig:prompt_layer_deberta_gpt2}
\end{figure}

\paragraph{Effect of prompt layer.} To enhance the reliability of the conclusion (i.e., the most intermediate layer of PTM is the optimal choice of the prompt layer) from Section~\ref{subsec:prompt_layer}, we also conduct the same experiments with Section~\ref{subsec:prompt_layer} on the other two PTMs that include DeBERTa\textsubscript{LARGE}~\citep{he2021deberta} and GPT2\textsubscript{LARGE}~\citep{radford2019gpt2} models. As shown in Figure~\ref{fig:prompt_layer_deberta_gpt2}, the most intermediate layer is also the optimal choice of the prompt layer on DeBERTa\textsubscript{LARGE} and GPT2\textsubscript{LARGE} models, especially for LPT w/ NPG. These results enhance the reliability of our conclusion that a better trade-off between performance and efficiency can be achieved by selecting the most intermediate layer of PTM as the prompt layer.

\begin{figure}[t]
    \centering
    \begin{subfigure}{0.49\linewidth}
    \centering
    \includegraphics[width=\linewidth]{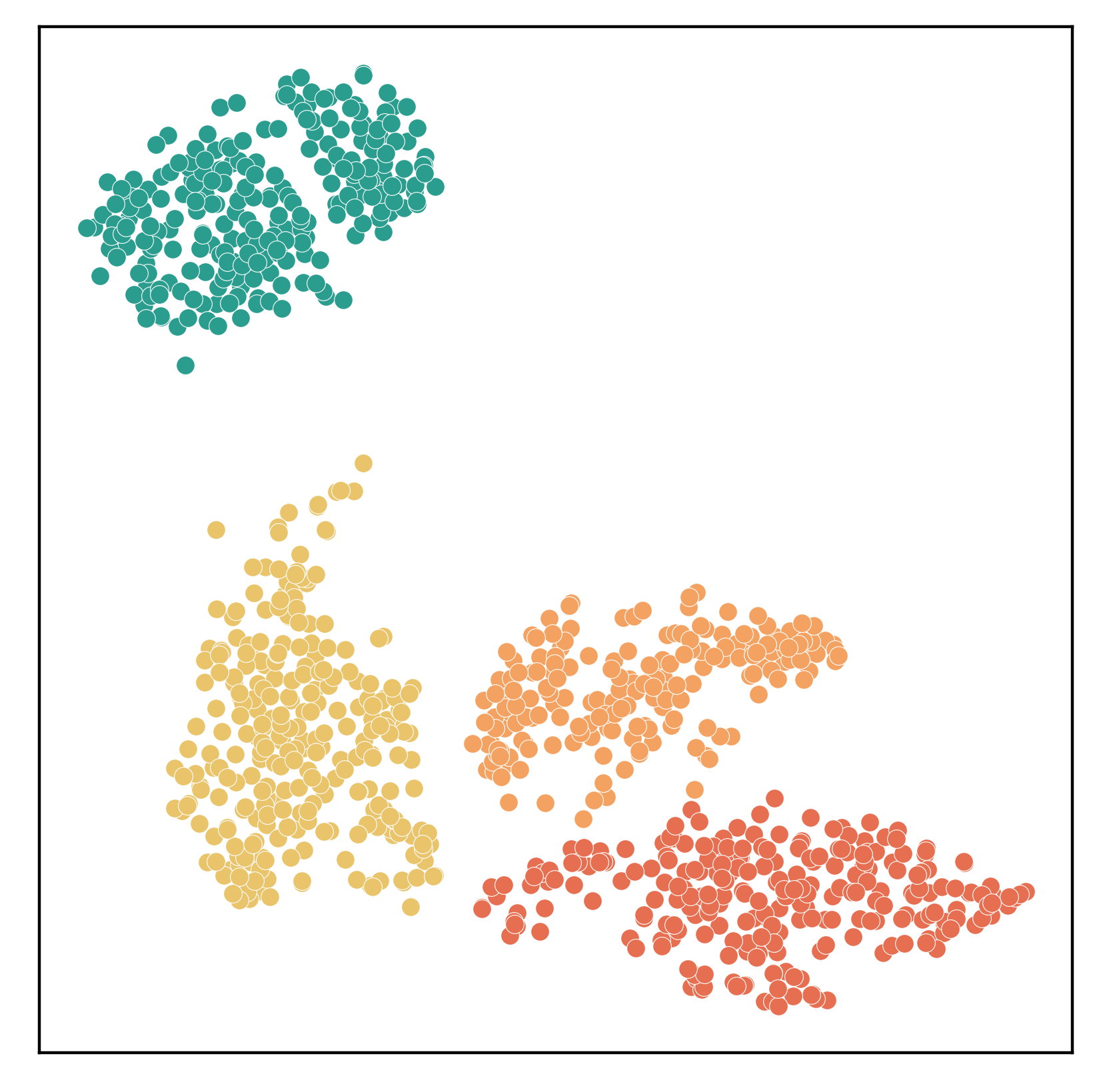}
    \end{subfigure}
    \begin{subfigure}{0.49\linewidth}
    \centering
    \includegraphics[width=\linewidth]{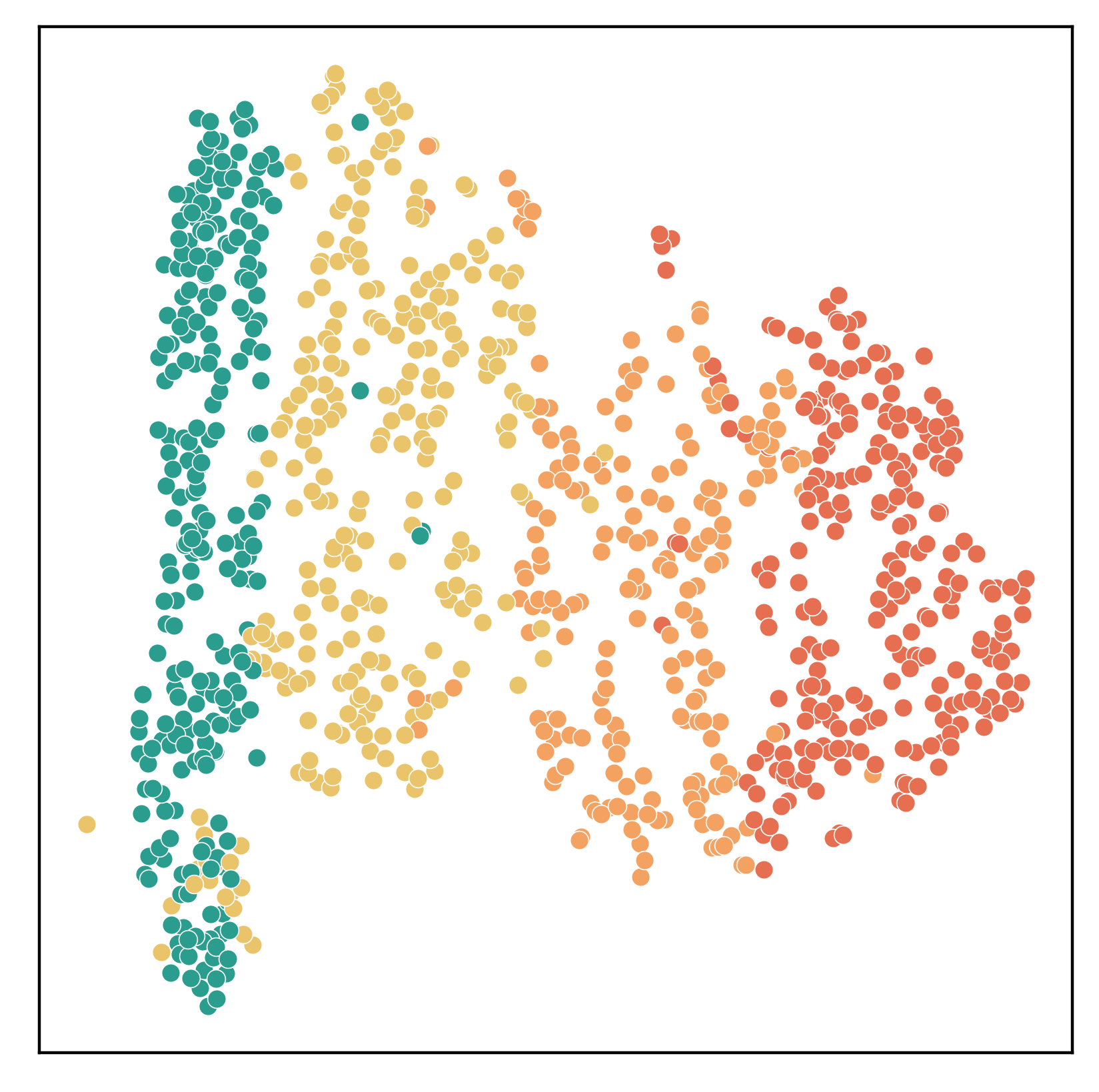}
    \end{subfigure}
    \caption{Sentence representation visualization (left) and instance-aware prompt visualization (right).}
    \label{fig:visualization}
\end{figure}

\paragraph{Visualization of instance-aware prompt.} We selected the subj dataset~\citep{pang2004subj} with 1000 development samples for this analysis. For the sake of simplification, we only visualize the instance-aware prompt of LPT w/ NPG method. As shown in Figure~\ref{fig:visualization}, we use the same color to mark the samples that their representations are close. We can clearly observe that our method can generate similar prompts for the instances with relatively similar sentence representation. On the contrary, the independent prompts of instances with quite different sentence representations are also quite different. The visualization result indicates that our method learns a special prompt for each instance and can be aware of the important information of the instance to drive PTMs better. 

\section{Conclusion}
In this paper, we explore why prompt tuning performs poorly and find there is a trade-off between the propagation distance from label signals to the inserted prompt and the influence of the prompt on model outputs. With this discovery, we present a more efficient and effective prompt tuning method LPT with late and instance-aware prompts. Experiment results in full-data and few-shot scenarios demonstrate LPT can achieve comparable or even better performance than state-of-the-art PETuning methods and full model tuning while having higher training speed and lower memory cost.

\section*{Limitations}
Although we showed that our proposed method can greatly improve performance and reduce training costs for diverse NLU tasks on three different PTMs (i.e., RoBERTa\textsubscript{LARGE}, DeBERTa\textsubscript{LARGE} and GPT2\textsubscript{LARGE}), the larger PTMs with billions of or more parameters and NLG tasks were not considered. But our main thought of using late and instance-aware prompt is simple and can be easily transferred to other backbone architectures and different types of tasks. It would be interesting to investigate if our findings hold for other backbone models and types of tasks. And we will explore it in future work.

\section*{Ethics Statement}
The finding and proposed method aims to improve prompt tuning in terms of training costs and performance. The used datasets are widely used in previous work and, to our knowledge, do not have any attached privacy or ethical issues.

\section*{Acknowledgements}
This work was supported by the National Key Research and Development Program of China (No.2020AAA0106700) and National Natural Science Foundation of China (No.62022027). 

\bibliography{emnlp2022}
\bibliographystyle{acl_natbib}

\newpage

\appendix

\begin{table*}[p]
\centering
\resizebox{0.98\linewidth}{!}{
\begin{tabular}{llcccccl}
\toprule
\textbf{Category}                         & \textbf{Datasets} & \textbf{$|$Train$|$} & \textbf{$|$Dev$|$}     & \textbf{$|$Test$|$}    & $\mathbf{|\mathcal{Y}|}$ & \textbf{Type}             & \textbf{Labels}                                        \\
\midrule
\multirow{5}{*}{Single-sentence} & SST-2    & 67349   & 872       & 1821      & 2   & sentiment        & positive, negative                            \\
                                 & MPQA     & 7606    & 1000      & 2000      & 2   & opinion polarity & positive, negative                            \\
                                 & MR       & 7662    & 1000      & 2000      & 2   & sentiment        & positive, negative                            \\
                                 & Subj     & 7000    & 1000      & 2000      & 2   & subjectivity     & subjective, objective                         \\
                                 & Trec     & 4952    & 500       & 500       & 6   & question cls.    & abbr., entity, description, human, loc., num. \\
\midrule
\multirow{5}{*}{Sentence-pair}   & MNLI     & 392702  & 19647 & 19643 & 3   & NLI              & entailment, neutral, contradiction            \\
                                 & MRPC     & 3668    & 408       & 1725      & 2   & paraphrase       & equivalent, not equivalent                    \\
                                 & QNLI     & 104743  & 5463      & 5463      & 2   & NLI              & entailment, not entailment                    \\
                                 & QQP      & 363846  & 40430     & 390965    & 2   & paraphrase       & equivalent, not equivalent                    \\
                                 & RTE      & 2490    & 277       & 3000      & 2   & NLI              & entailment, not entailment \\
\bottomrule
\end{tabular}
}
\caption{The statistics of datasets evaluated in this work. For MNLI task, the number of samples in development and test sets is summed by matched and mismatched samples. $|\mathcal{Y}|$ is the number for classes.}
\label{tab:data_statistics}
\end{table*}

\begin{table*}
\resizebox{\linewidth}{!}{
\begin{tabular}{lccccccl}
\toprule
\multirow{2}{*}{\textbf{Hyperparameter}} & \multicolumn{2}{c}{\textbf{RoBERTa}}                 & \multicolumn{2}{c}{\textbf{DeBERTa}}                 & \multicolumn{2}{c}{\textbf{GPT2}}                    &  \\
                                & \textbf{Full-data}            & \textbf{Few-shot}             & \textbf{Full-data}            & \textbf{Few-shot}             & \textbf{Full-data}            & \textbf{Few-shot}             &  \\
\midrule
\#Layers                        & 24                   & 24                   & 24                   & 24                   & 36                   & 36                 &  \\
Hidden size                     & 1024                 & 1024                 & 1024                 & 1024                 & 1280                 & 1280                 &  \\
Dropout rate                    & 0.1                  & 0.1                  & 0.1                  & 0.1                  & 0.1                  & 0.1                  &  \\
Peak learning rate              & 5e-4$-$1e-2 & 5e-4$-$1e-2 & 5e-4$-$1e-2 & 5e-4$-$1e-2 & 5e-4$-$1e-2 & 5e-4$-$1e-2 &  \\
Warmup type                     & linearly decayed     & linearly decayed     & linearly decayed     & linearly decayed     & linearly decayed     & linearly decayed     &  \\
Warmup rate                     & \{0, 0.06\}          & \{0, 0.06\}          & \{0, 0.06\}          & \{0, 0.06\}          & \{0, 0.06\}          & \{0, 0.06\}          &  \\
Batch size                      & \{16, 32\}           & \{8, 16, 32\}           & \{16, 32\}           & \{8, 16, 32\}           & \{8, 16\}         & \{4, 8, 16\}         &  \\
Weight decay                    & 0.1                  & 0.1                  & 0.1                  & 0.1                  & 0.1                  & 0.1                  &  \\
Training step                   & $-$                   & 1000                 & $-$                   & 1000                 & $-$                   & 1000                 &  \\
Training epoch                  & 10                   & $-$                   & 10                   & $-$                   & 10                   & $-$                   &  \\
AdamW $\beta_1$\                & 0.9                  & 0.9                  & 0.9                  & 0.9                  & 0.9                  & 0.9                  &  \\
AdamW $\beta_2$\                & 0.999                & 0.999                & 0.999                & 0.999                & 0.999                & 0.999                &  \\
AdamW $\epsilon$                & 1e-8                 & 1e-8                 & 1e-8                 & 1e-8                 & 1e-8                 & 1e-8                 & \\
\bottomrule
\end{tabular}}
\caption{The search space for each hyperparameter considered in our method.}
\label{tab:hyperparam}
\end{table*}

\begin{table*}
\resizebox{0.98\linewidth}{!}{
\begin{tabular}{lll}
\toprule
\textbf{Task} & \textbf{Template} & \textbf{Label words}                              \\
\midrule
SST-2                          & $\langle S_1 \rangle$ It was \texttt{[MASK]} .               & positive: great, negative: terrible                                \\
MPQA                           & $\langle S_1 \rangle$ It was \texttt{[MASK]} .              & positive: great, negative: terrible                                \\
MR                             & $\langle S_1 \rangle$ It was \texttt{[MASK]} .              & positive: great, negative: terrible                                \\
Subj                           & $\langle S_1 \rangle$ It was \texttt{[MASK]} .              & subjective: subjective, objective: objective                       \\
TREC                           & \texttt{[MASK]} : $\langle S_1 \rangle$                 & abbreviation: Expression, entity: Entity, description: Description \\
                               &                                    & human: Human, location: Location, numeric: Number                  \\
\midrule
MNLI                           & $\langle S_1 \rangle$ ? \texttt{[MASK]} , $\langle S_2 \rangle$       & entailment: Yes, netural: Maybe, contradiction: No                 \\
MRPC                           & $\langle S_1 \rangle$ \texttt{[MASK]} , $\langle S_2 \rangle$         & equivalent: Yes, not equivalent: No                                \\
QNLI                           & $\langle S_1 \rangle$ ? \texttt{[MASK]} , $\langle S_2 \rangle$       & entailment: Yes, not entailment: No                                \\
QQP                            & $\langle S_1 \rangle$ \texttt{[MASK]} , $\langle S_2 \rangle$         & equivalent: Yes, not equivalent: No                                \\
RTE                            & $\langle S_1 \rangle$ ? \texttt{[MASK]} , $\langle S_2 \rangle$       & entailment: Yes, not entailment: No \\
\bottomrule
\end{tabular}
}
\caption{The manual templates and label words used on RoBERTa and DeBERTa models. }
\label{tab:template_roberta}
\end{table*}

\begin{table*}
\resizebox{0.98\linewidth}{!}{
\begin{tabular}{lll}
\toprule
\textbf{Task} & \textbf{Template} & \textbf{Label words}                              \\
\midrule
Subj & $\langle S_1 \rangle$ It was \texttt{[MASK]} .              & subjective: subjective, objective: objective                       \\
TREC & \texttt{[MASK]} : $\langle S_1 \rangle$                     & abbreviation: Expression, entity: Entity, description: Description \\
     &                                        & human: Human, location: Location, numeric: Number                  \\
\midrule
MRPC & $\langle S_1 \rangle$  $\langle S_2 \rangle$ They are \texttt{[MASK]} . & equivalent: Yes, not equivalent: No                                \\
RTE  & $\langle S_1 \rangle$  $\langle S_2 \rangle$ They are \texttt{[MASK]} . & entailment: Yes, not entailment: No  \\
\bottomrule
\end{tabular}
}
\caption{The manual templates and label words used on GPT2 model. }
\label{tab:template_gpt2}
\end{table*}

\section{Details for Mutual Information Estimation}
\label{sec:mutual_information}
Because the mutual information cannot be be calculated directly, we estimate it by training a new classifier using the hidden states $\mathbf{h}$ as inputs and the original labels of inputs as outputs. Then, we estimate $I(\mathbf{h}, y)$ using the performance achieved by the classifier. Since $I(\mathbf{h}, y) = H(y) - H(y|\mathbf{h}) = H(y) - \mathbb{E}_{(\mathbf{h}, y)}[-\rm log$ $p(y|\mathbf{h})]$~\citep{wang2021lsl}, we can train a new classifier $q_{\psi}(y|\mathbf{h})$ to approximate $p(y|\mathbf{h})$, such that we have $I(\mathbf{h}, y) \approx \rm{max}_{\psi} \{\mathit{H(y)} - \frac{1}{\mathit{N}}[\mathit{\sum}_{\mathit{i}=1}^{\mathit{N}}-\rm log$ $q_{\psi}(y_i|\mathbf{h}_i)]\}$. Because $H(y)$ is a constant, we are going to ignore it here. Based on the above conditions, we can use the loss of $q_{\psi}(y|\mathbf{h})$ (i.e., $- \frac{1}{\mathit{N}}[\mathit{\sum}_{\mathit{i}=1}^{\mathit{N}}-\rm log$ $q_{\psi}(y_i|\mathbf{h}_i)]$) as the estimate of $I(\mathbf{h}, y)$. Further simplification, we use the performance of this new classifier to estimate mutual information $I(\mathbf{h}, y)$. Because RoBERTa\textsubscript{LARGE}~\citep{liu2019roberta} has 24 layers totally except embedding layer, we can obtain 24 hidden states for each input. Hence, we need to train 24 new classifiers for each method. To speed up the training process, we use a 6-layer RoBERTa\textsubscript{LARGE} as $q_{\psi}$.

\section{Datasets}
\label{sec:dataset_appendix}
For SST-2~\citep{socher2013sst-2}, MNLI~\citep{williams2018mnli}, MRPC~\citep{dolan2005mrpc}, QNLI~\citep{rajpurkar2016qnli}, QQP\footnote{\url{https://www.quora.com/q/quoradata/}} and RTE~\citep{dagan2005rte} datasets which are from GLUE benchmark~\citep{wang2019glue}, we use their original data splits. For 4 other datasets, we select a certain number of samples from the training set as the development set, and the number of samples for each label is determined according to its proportion in the original training set. The dataset statistics after split are shown in Table~\ref{tab:data_statistics}

\section{Implementation Details}
\label{sec:hyper&template}
 The search space of hyperparameters considered in this paper is shown in Table~\ref{tab:hyperparam}. As an additional note, we use the same number of training epochs or steps for all the methods. For adapter-based tuning methods, we set the down-projection size $m$ to 16. We set the prompt length to 20 for prompt tuning~\citep{lester2021pt} and P-tuning v2~\citep{Liu2022ptuningv2}, and 5 for S-IDPG-PHM~\citep{Wu2022idpg} and LPT w/ NPG. For LPT w/ MPPG and LPT w/ APPG, due to the number of tunable parameters being invariable with prompt length changes, we also search the prompt length in the range of \{10, 15, 20\} for them. Besides, we set the down-projection size $m$ of S-IDPG-PHM and LPT to 256 and 128, respectively. The hyperparameter $r$ and $\alpha$ in LoRA are set to 8 and 16 on RoBERTa\textsubscript{LARGE}, 4 and 32 on GPT2\textsubscript{LARGE}. For the batch size of GPT2 model listed in Table~\ref{tab:hyperparam}, it refers to the number of samples in a single forward pass. Due to the large scale of GPT2\textsubscript{LARGE}, we use \textit{gradient accumulation} technique to avoid out-of-memory, and the accumulation step is 2 or 4. We use AdamW optimizer~\citep{loshchilov2019adamw} for all the methods in this work. We use Pytorch~\citep{paszke2019pytorch} and HuggingFace's Transformers~\citep{wolf2020huggingface} libraries to implement all the methods in this work. All experiments are conducted on 8 NVIDIA GTX 3090 GPUs.

We follow \citet{gao2021lmbff} and show the used manual templates and label words in Table~\ref{tab:template_roberta} and Table~\ref{tab:template_gpt2}, respectively. Note that, since the vocabulary of the GPT2 model doesn't have the \texttt{[MASK]} token, we justly use it to represent the positions that are needed to predict.

\begin{table}
\centering
\resizebox{\linewidth}{!}{
\begin{tabular}{ccccc}
\toprule
\multirow{2}{*}{\textbf{Method}} & \multirow{2}{*}{\textbf{Accuracy}} & \textbf{Tunable}    & \textbf{Training Speed} & \textbf{Memory Cost} \\
                        &                           & \textbf{Parameters} & tokens/ms (\textcolor{blue}{$\uparrow$})      & GB (\textcolor{blue}{$\downarrow$})          \\
\midrule
Model Tuning            & 52.0 (1.9)                & \cellcolor[gray]{0.55}355M       & 11.6           & 23.5        \\
\hdashline
Prompt Tuning           & 58.2 (1.7)                & \cellcolor[gray]{0.95}21K        & 16.9 (\textcolor{blue}{1.5$\times$})    & 17.8 (\textcolor{blue}{24.3\%}) \\
\midrule
\multicolumn{5}{c}{\textit{LPT w/ NPG}} \\                \midrule
PL = 7                    & \underline{63.2} (3.3)                & \cellcolor[gray]{0.75}792K       & 18.5 (\textcolor{blue}{1.6$\times$})    & 13.4 (\textcolor{blue}{43.0\%}) \\
PL = 13                   & \textbf{69.5} (3.1)                & \cellcolor[gray]{0.75}792K       & 23.2 (\textcolor{blue}{2.0$\times$})    & 10.1 (\textcolor{blue}{56.6\%}) \\
PL = 19                   & 62.6 (3.3)                & \cellcolor[gray]{0.75}792K       & 28.5 (\textcolor{blue}{2.5$\times$})    & \; 6.7 (\textcolor{blue}{71.5\%})  \\
\midrule
\multicolumn{5}{c}{\textit{LPT w/ MPPG}} \\
\midrule
PL = 7                    & 59.9 (4.4)                & \cellcolor[gray]{0.85}263K       & 19.8 (\textcolor{blue}{1.7$\times$})    & 14.3 (\textcolor{blue}{39.1\%}) \\
PL = 13                   & 62.4 (3.1)                & \cellcolor[gray]{0.85}263K       & 23.4 (\textcolor{blue}{2.0$\times$})   & 10.6 (\textcolor{blue}{54.9\%}) \\
PL = 19                   & 58.8 (1.5)                & \cellcolor[gray]{0.85}263K       & 28.8 (\textcolor{blue}{2.5$\times$})    & \; 7.0 (\textcolor{blue}{70.2\%})  \\
\midrule
\multicolumn{5}{c}{\textit{LPT w/ APPG}}  \\
\midrule
PL = 7                    & 58.6 (2.3)                & \cellcolor[gray]{0.85}263K       & 19.8 (\textcolor{blue}{1.7$\times$})    & 14.3 (\textcolor{blue}{39.1\%}) \\
PL = 13                   & 63.0 (2.2)                & \cellcolor[gray]{0.85}263K       & 23.4 (\textcolor{blue}{2.0$\times$})    & 10.6 (\textcolor{blue}{54.9\%}) \\
PL = 19                   & 60.1 (2.2)                & \cellcolor[gray]{0.85}263K       & 28.8 (\textcolor{blue}{2.5$\times$})    & \; 7.0 (\textcolor{blue}{70.2\%})  \\
\bottomrule
\end{tabular}
}
\caption{Trade-off between performance and training efficiency. 'PL' denotes the prompt layer. \textbf{Bold} and \underline{Underline} marks the best and the second best results, respectively. All methods are evaluated on RTE dataset using RoBERTa\textsubscript{LARGE} model.}
\label{tab:efficiency_pl}
\end{table}

\section{Efficiency Evaluation on Different Prompt Layers.}
\label{sec:efficiency_pl}
We select the prompt layer in the range of $\{7, 13, 19\}$ to explore the influence from different prompt layers for the trade-off between efficiency and performance. The experiment settings are consistent with those described in Section~\ref{sec:efficiency}. Table~\ref{tab:efficiency_pl} shows the performance, the number of tunable parameters, training speed, and memory cost for LPT with three different prompt layers. When the prompt layer is the 13th layer, both performance and training efficiency are better than when it is the 7th layer. When the prompt layer is the 19th layer, the efficiency is further improved while the performance degrades a lot.

\end{document}